\documentclass[a4paper,fleqn]{cas-dc}

\usepackage{paralist, tabularx}
\usepackage{subcaption}
\usepackage{algorithm2e}
\usepackage[numbers]{natbib}
\usepackage{tabularx}

\usepackage[utf8]{inputenc}
\usepackage[T1]{fontenc}
\usepackage{bbm}
\usepackage{enumitem}
\setlist{nolistsep}
\usepackage{amsmath,amsfonts}

\def\tsc#1{\csdef{#1}{\textsc{\lowercase{#1}}\xspace}}
\tsc{WGM}
\tsc{QE}
\tsc{EP}
\tsc{PMS}
\tsc{BEC}
\tsc{DE}

\begin{document}
\let\WriteBookmarks\relax
\def\floatpagepagefraction{1}
\def\textpagefraction{.001}



\title [mode = title]{\textcolor{black}{Ensemble quantile-based deep learning framework for   streamflow and flood prediction in  Australian catchments}}                       

\author[3]{Rohitash Chandra \corref{cor1}} 
\ead{rohitash.chandra@unsw.edu.au} 

 \author[3]{Arpit Kapoor} %
\author[1,4]{
Siddharth Khedkar}    
 
\author[3]{Jim Ng}    

\author[2]{R. Willem Vervoort} 
\affiliation[3]{Transitional Artificial Intelligence Research Group, School of Mathematics and Statistics, UNSW Sydney, Sydney, Australia}

\affiliation[2]{ARC ITTC in Data Analytics for Resources and Environments, The University of Sydney,   Sydney, Australia}

\affiliation[1]{Birla Institute of Technology and Sciences Pilani, Rajasthan, India}

\affiliation[4]{Centre for Artificial Intelligence and Innovation, Pingla Institute, Sydney, Australia}

\cortext[cor1]{Corresponding author}



\begin{abstract}
In recent years, climate extremes such as floods have created significant environmental and economic hazards for Australia.  Deep learning methods have been promising for predicting extreme climate events; however, large flooding events present a critical challenge due to factors such as model calibration and missing data.   We present an ensemble quantile-based deep learning framework that addresses large-scale streamflow forecasts using quantile regression for uncertainty projections in prediction. We evaluate selected univariate and multivariate deep learning models and catchment strategies. We implement a multistep time-series prediction model \textcolor{black}{using the CAMELS dataset for selected catchments across Australia. The ensemble model employs a set of quantile deep learning models for streamflow determined by historical streamflow data. We utilise the streamflow prediction and obtain flood probability using flood frequency analysis and compare it with historical flooding events for selected catchments.} Our results demonstrate notable efficacy and uncertainties in streamflow forecasts with varied catchment properties. \textcolor{black}{Our flood probability estimates show good accuracy in capturing the historical floods from the selected catchments.  This underscores the potential for our deep learning framework to revolutionise flood forecasting across diverse regions and be implemented  as an early warning system.}

 
\end{abstract}



\begin{keywords} 
Streamflow \sep floods \sep climate extremes \sep deep learning \sep quantile regression
\end{keywords}

\maketitle
\section{Introduction}

Flooding is among the most devastating and costly natural disasters in Australia in terms of economic loss \cite{smith1994flood,halgamuge2017analysis}, damage to natural habitats and environment, and loss of human and animal lives \cite{IPCCwebsite,macmahon2015connecting}.  Fernandez et al. \cite{fernandez2015flooding} highlighted the psychological impact of floods and suggested that floods can cause post-traumatic stress disorder, depression and anxiety, and long-term mental health implications for flood victims. It is the rare and extreme floods that have the most far-reaching consequences for both humans and the environment.  Between 1998 and 2022, multiple extreme floods occurred in New South Wales (NSW) and Queensland, causing between \$900 million and \$3.4 billion losses. As well as causing fatalities, economic impacts continued for years afterwards. 

Machine learning methods have been increasingly used to tackle time series prediction problems for a wide range of domains \cite{sezer2020financial,han2019review,lim2021time,langkvist2014review} and are extensively used in hydrology \cite{lange2020machine,zounemat2021ensemble}.  Machine learning methods can forecast streamflow at various lead times, and show promise for hydrological forecasts of streamflow and floods \cite{kicsi2011combined, campolo1999river, hsu1995artificial, zhang2009estimating, lima2016forecasting, zhou2019explore}. 
Hence, the literature motivates the use of machine learning models for extreme climate events such as floods in Australia.

 A major challenge in hydrology is the 'regional modelling problem', which is about the usage of either a single model or a set of models to provide spatially continuous hydrological simulations across large areas (e.g., regional, continental, global). Regional models must learn and encode the similarities and differences of various catchment characteristics (such as topology, geometry, climate, ecology, etc.) and use them to meaningfully predict hydrological behaviour. Often this is done using ancillary data (e.g. soil maps, remote sensing, digital elevation maps, etc.) to map the similarity between catchments \cite{razavi2013streamflow}. \textcolor{black}{Kratzert et al. \cite{kratzert2019benchmarking} showed how a Long Short-Term Memory (LSTM) \cite{hochreiter1997long,greff2016lstm} deep learning model could be used to develop a single (regional) hydrologic model for hundreds of sites, concatenating the static data to the time-series data as inputs. LSTM models have outperformed traditional hydrological models for streamflow prediction in ungauged basins \cite{arsenault2023hess}. The compression of the static information used in hydrologic modelling can improve the integration of the static information with the time series data \cite{shalev2019accurate} using ensemble learning.}
 
 Ensemble learning models use multiple machine learning models and combine their predictions for improved performance \cite{sagi2018ensemble,khan2023review}.   Leontjeva et al. \cite{leontjeva2016combining} showed that an LSTM model can extract information from temporal sequences, which can then be combined with ensemble methods like Extreme Value Machines and Random Forests to utilise static data. Miebs et al. \cite{miebs2020efficient} demonstrated that an ensemble  of Recurrent Neural Networks (RNNs)  and Dense Networks could learn the temporal data with static supplementary information. Therefore, an ensemble    learning models can enable a regional approach to potentially outperform a single-site approach calibrated per basin, which often suffers from inherent data shortage.

Forecasting time series with extremes is very difficult, as the extreme values are often treated as outliers to improve the generalisation of the model. Most research on extreme values focuses on using Extreme Value Theory (EVT) \cite{de2007extreme} for detecting that an extreme event is present, rather than being able to predict the magnitude of the extreme event.   Galib et al. \cite{galib2022deepextrema}  used simple neural networks to predict the parameters of an extreme value distribution, which consequently could predict the highest value within a specific time window of the time-series data.

Quantile Regression models the conditional median (or other quantiles) of the \textit{response variable}, whereas Linear Regression  estimates the conditional mean. Similar to linear regression, a quantile regression model training procedure minimises the mean absolute error to estimate conditional quantiles of the response variable \cite{koenker1978regression}.  The quantile regression model is more robust to outliers in the response measurements and hence applicable to forecasting extremes \cite{portnoy1999extreme,cai2013extreme}. Quantile regression has been combined with machine learning models for several applications.  Taylor et al. \cite{taylor2000quantile} introduced the quantile regression neural network, which offered a practical and flexible implementation compared to past work that involved linear or simple parametric non-linear models.  Wang et al. \cite{wang2019probabilistic} combined quantile regression with an LSTM model for load forecasting, and  Pasche et al. \cite{pasche2022neural} combined extreme value theory with quantile regression for a novel model which was applied to flood risk forecasting. Quantile regression found multiple applications in hydrology, such as in down-scaling precipitation \cite{bremnes2004probabilistic} and climate characteristics \cite{timofeev2010using}. In the case of flood forecasting, Weerts et al.  \cite{weerts2011estimation} applied quantile regression to assess the predictive uncertainty of rainfall-runoff and hydraulic forecasts.  Cai and Reeve \cite{cai2013extreme} addressed extreme and non-extreme value prediction via parametric quantile function models, showcasing differences among statistical models commonly used in extreme value prediction.

 The domain of extreme flood prediction, particularly through the integration of deep learning with quantile regression, presents a unique opportunity to enhance the accuracy of flood forecasts. Despite significant advancements in forecasting streamflow and rainfall-runoff modelling, there is a critical gap concerning the prediction of extreme flood events that can enable the prediction of floods. This gap could be addressed by combining deep learning with quantile regression, which so far has not been used widely. This will specifically allow for assessing predictive uncertainty in hydrological forecasts.  Furthermore, integration of quantile deep learning with ensemble learning \cite{ganaie2022ensemble,yang2023survey} can further address the problem of model uncertainty  quantification and enable uncertainty projection in forecasts.
 
 In this study, we present an ensemble quantile deep learning framework that addresses large-scale streamflow and flood probability forecasts with uncertainty quantification in predictions.
Our framework employs a multivariate and multi-step time-series prediction approach to forecast streamflow for multiple days ahead using the \textit{catchment attributes and meteorological} (CAMELS) dataset \cite{fowler2021camels} for selected catchments across  Australian states.  \textcolor{black}{We evaluate selected univariate and multivariate deep learning models and catchment strategies, including individual and 
ensemble learning. We also employ static information to enrich the time-series information, allowing regional modelling across catchments. 
The ensemble model employs a set of quantile deep learning models for streamflow determined by historical streamflow data. We utilise the streamflow prediction and obtain flood probability using flood frequency analysis and compare it with historical flooding events for selected catchments.} We compare the performance of selected deep learning models for selected basins across the Australian states.  \textcolor{black}{A major contribution of our study is the development of a quantile deep learning model within an ensemble model, which has major potential for hydrology.} 
We provide open-source code along with data so that the methodology can be extended to other catchments in other regions.

The rest of the paper is organised as follows.  In Section 2, we provide an overview of related topics, and in Section 3 we present our methodology. Section 4 provides the results, followed by a discussion in Section 5, and a conclusion in Section 6. 

\section{Background and Related Work}

\subsection{Hydrological models}\label{Static}

Surface water flooding is a result of accumulated rainfall covering a land surface since it cannot drain to natural or man-made drainage systems or watercourses due to them being at full capacity and overflowing \cite{speight2019towards}. \textcolor{black}{In short-duration intense rainfall,} flash flooding can arise from rapidly developing convective rainfall systems, which can be particularly dangerous in urban areas \cite{adams2016flood}. The prediction of flash floods for small to medium catchments remains a great challenge, as flash floods aren't always caused by known meteorological phenomena \cite{nguyen2020influence}. Some factors that cause flash floods may include how quickly the storm is moving, the porosity of the soil, the steepness of the terrain, etc. 
Climate change is a big driver of extreme weather events that cause floods \cite{CCwebsite}, and in particular rainfall events have been increasing in intensity \cite{wasko2023climatchangeeffects}.

The most common hydrological forecasting models are combined hydrological/hydrodynamic and meteorological models \cite{bartholmes2005coupling}. 
Hydrological models mostly simulate rainfall-runoff relationships, while hydraulic models deal with channel flow over surfaces. Meteorological rainfall forecast models are needed, as ultimately rainfall drives floods, but these need to quantify uncertainty \cite{speight2021operational}. Several major  sources of uncertainty  limit forecasting ability  \cite{krzysztofowicz2001forecastuncertainty} and the general approach to managing uncertainty has been to use ensemble forecasts. For example, Emerton et al. \cite{emerton2016continental} used ensemble streamflow forecasting and featured several forecast scenarios to explicitly forecast uncertainties. At a much larger scale, Alfieri et al. \cite{alfieri2013glofas} presented a Global Flood Awareness System (GloFAS) for detecting the probability of streamflow given warning thresholds by using ensemble models. However, Cloke et al. \cite{cloke2009ensemble} outlined that ensemble models can be limited by computational speed and complexity. Ultimately, hydrological models are constrained by the amount of information available, such as the input of meteorological variables (e.g. flow, water level, rainfall) and the need to parameterise geological, topographic and soil variables \cite{merwade2008gis}.


As mentioned, most hydrological models are calibrated for a single catchment, and the challenge is to extrapolate to different catchments across a region. In regionalisation approaches, model-dependent (parametric) and model-independent (non-parametric) approaches are the two primary strategies \cite{razavi2013streamflow}. The model-dependent approaches assume a pre-defined hydrological model across all catchments, wherein the model parameters are estimated from the data. Different approaches can be used for regionalisation, focusing on either spatial proximity or hydrological similarity. \textcolor{black}{In a study by Oudin et al.  \cite{oudin2008regionalisation}, spatial proximity gave the best performance across the selected catchments in France; however, the models calibrated to single catchments outperformed all the other approaches.} In a recent uncertainty-focused approach, Prieto et al. \citep{prieto2022bayesianregionalisation} estimated not only the model parameters, but different model structures in a regionalisation approach. 

Conversely, model-independent methods, non-parametric in nature, require no prior assumptions and are driven entirely by the data \cite{arsenault2023hess}. It identifies the mapping between ancillary data, catchment attributes and other inputs to predict streamflow. The most common of these are machine learning models (such as neural networks) \textcolor{black}{that have been most effective with large amounts of data and enable model knowledge to be transferred between sites \cite{harter2005role}. Several studies  \cite{kratzert2018rainfall, kratzert2019benchmarking,ni2020streamflow} have shown that LSTM models, in particular, can outperform conceptual hydrological models,} even if they are calibrated individually for each catchment in a gauged setting \cite{kratzert2018rainfall}.

\subsection{Deep learning  models}\label{RNN}

 The success of  RNNs has been prominent in language models \cite{elman1988learning} and modelling temporal sequences \cite{fang2021survey}, including multistep-ahead prediction \cite{chandra2021evaluation}.  The major categories of time series forecasting (prediction) include one-step, multi-step and multivariate time series prediction. The one-step-ahead prediction refers to models that have a prediction horizon (step) of a single time unit, whereas multi-step predictions refer to models that have a prediction horizon of multiple time units. Developing models that maintain good prediction accuracy as the prediction horizon increases is challenging \cite{taieb2015bias,chandra2021evaluation}. \textcolor{black}{In certain cases, such as extreme value forecasting in climate-related problems, and cryptocurrency forecasting, the chaotic nature of the data and the missing features make a significant difference in model performance. RNN models had the problem of learning long-term dependencies in temporal data, due to the limited  "memory" in capturing the autocorrelation present in long-term data streams  \cite{bengio1994learning}. The LSTM network \cite{hochreiter1997long} addressed this limitation} through the inclusion of internal memory cells that control the flow of information in the model.  This has improved Natural Language Processing (NLP) \cite{karpathy2015visualizing} and Machine Translation \cite{sutskever2014sequence} which involve longer-term dependencies  in text data,    i.e a paragraph in a text can be the entire chapter that needs to be translated.  LTSM models can capture memory and therefore autocorrelation, as well as linear and non-linear behaviour in hydrological systems.  LSTM models can learn long-term dependencies for temporal sequences and exhibit better performance for flood forecasting \cite{le2019application} when compared to simple neural networks.   

Bidirectional RNNs \cite{schuster1997bidirectional} further addressed the shortcomings of earlier counterparts, where only direct previous context states are used for determining future states. The bidirectional architecture consists of placing two independent RNN layers together to allow information to be processed in both directions, at every time step \cite{schuster1997bidirectional}. Accordingly, information is transferred from past to future and from future to past time-steps, and bidirectional LSTM network implementation enhances conventional LSTM network \cite{graves2005framewise}.  The encoder-decoder LSTM \cite{sutskever2014sequence} is an RNN originally designed to process sequence-to-sequence problems in text data processing, where the number of items in output and output sequences varies. In recent studies, the encoder-decoder LSTM has shown to be one of the best models, overtaking LSTM, CNN, and BD-LSTM for multi-step time series prediction problems \cite{chandra2021evaluation,wu2024review}.

 CNNs \cite{wang2017deep,shi2015convolutional}  have been designed to learn spatial hierarchies of features through multiple building blocks of convolution, pooling and fully connected layers. The convolutional layers were originally designed for image classification purposes \cite{krizhevsky2012imagenet}, but can also be used for feature extraction from temporal sequences, including time series prediction \cite{chandra2021evaluation}.  For example, the CNN has been used for precipitation downscaling \cite{vandal2017deepsd} and a wide time of time series prediction tasks \cite{wu2024review}.   In particular, Chandra et al. \cite{chandra2021evaluation} showed that CNN is competitive with LSTM  for multi-step ahead time series prediction. \textcolor{black}{Furthermore, Hussain et al. \cite{hussain2020deep} implemented a one-dimensional (1D) CNN  for one-step-ahead streamflow forecasting for three-time horizons (daily, weekly and monthly) in Gilgit River of Pakistan. Khosravi et al. \cite{khosravi2022using} used CNN with meta-heuristic optimisation and reported superior predictive performance in daily stream-flow forecasting compared to related models.}

\subsection{Ensemble learning}

Ensemble learning models combine multiple machine learning methods as base models to produce an optimal predictive model \cite{ganaie2022ensemble,dong2020survey}. Some of the prominent ensemble learning models are Bagging and Boosting methods via implementations such as Random Forests and \textit{AdaBoost} \cite{sagi2018ensemble}, respectively.  \textcolor{black}{The combination of ensemble learning with deep learning models have become prominent in the last decade \cite{ganaie2022ensemble}. Ensemble models have been applied extensively in hydrology\cite{zounemat2021ensemble}, with a focus on how to construct the ensemble given multiple channels of flow data using averaging   \cite{diks2010comparison}, bootstrapping \cite{sharma2009bootstrap}, and stacking \cite{tyralis2021super,li2020multi} methods.} \textcolor{black}{Initially,  Cannon et al. \cite{cannon2002downscaling}  investigated how climate change impacted streamflow using ensemble neural networks and multiple linear regression methods.   Diks an Vrugt \cite{diks2010comparison} compared different model averaging techniques and indicated that the Granger-Ramanathan averaging method provided the best accuracy for streamflow forecasting. In contrast, Sharma et al. \cite{sharma2009bootstrap} demonstrated that a hybrid wavelet bootstrapped neural network provided better predictions than conventional machine learning models in forecasting daily river discharge. Tyralis et al. \cite{tyralis2021super}  forecasted short-term daily streamflow by stacking ten different machine learning models. Furthermore, Li et al. \cite{li2020multi}   integrated stacked models including Support Vector Machines, Elastic-net Regression and  Extreme Gradient Boosting} (XGBoost)  to forecast midterm streamflow. However, bagging has been the predominant ensemble model until 2010 in hydrology and more recently, boosting and stacking have dominated due to improved performance \cite{zounemat2021ensemble}.

\subsection{Quantile Regression}

Koenker et al. \cite{koenker1978regression} introduced Quantile Regression for estimating point values of individual quantiles directly using regression models.  Since then, quantile regression has found multiple applications in hydrology, such as precipitation downscaling \cite{bremnes2004probabilistic} and climate characteristics \cite{timofeev2010using}. Friederichs et al. \cite{friederichs2007statistical} developed a censored quantile regression model to predict mixed discrete-continuous variables such as precipitation amounts, wind speeds, or pollutant concentrations. In the case of flood forecasting, Weerts et al. \cite{weerts2011estimation} applied quantile regression in assessing predictive uncertainty of rainfall-runoff and hydraulic forecasts.

The quantile regression model estimates the conditional median of the response variable, as compared to regular regression, which estimates the conditional mean through the least squares method. This is achieved through minimising the mean absolute error, and by applying asymmetric weights to positive/negative errors, the conditional quantiles of the response variable can be estimated instead \cite{koenker1978regression}. Suppose $Y$ is the real-valued random variable for the response, then $F_Y(y) = P(Y \leq y)$ represents the cumulative distribution function. The $\tau^{th}$ quantile for $\tau \in (0,1)$ may be given by: $$q_Y(t) = F_Y^{-1}(\tau) = inf\{y: F_Y(y) \geq \tau\}$$



 We define the tilted absolute value function as $\rho_{\tau}(m)$ for a particular quantile $\tau$:
\begin{equation}
  \rho_{\tau}(m) = m(\tau-\mathbbm{1}_{m<0})
\label{eqn:qnt}
\end{equation}

where $\mathbbm{1}$ is the indicator function. The specific quantile can be found by minimising the expected loss of $Y-u$, with respect to $u$:

\begin{equation}
q_Y(\tau) = \underset{u}{\arg\min}\text{ } \mathbb{E}[\rho_{\tau}(Y-u)]
\label{eqn:loss}
\end{equation}



\textcolor{black}{Quantile regression has been combined with machine learning models for several applications. Taylor et al. \cite{taylor2000quantile} introduced the quantile regression neural network, which offered a practical and flexible implementation compared to past work that involved linear or simple parametric non-linear models. Wang et al. \cite{wang2019probabilistic} combined quantile regression with an LSTM model for load forecasting, and Pasche et al. \cite{pasche2022neural} combined extreme value theory with quantile regression for a novel model which was applied to flood risk forecasting. These combinations motivate our work, which focuses on developing a quantile deep learning model for streamflow forecasting. }

\section{Data and Methodology}

\subsection{CAMELS dataset} \label{CAMELS}

\begin{figure}
\centering
\includegraphics[scale=0.5]{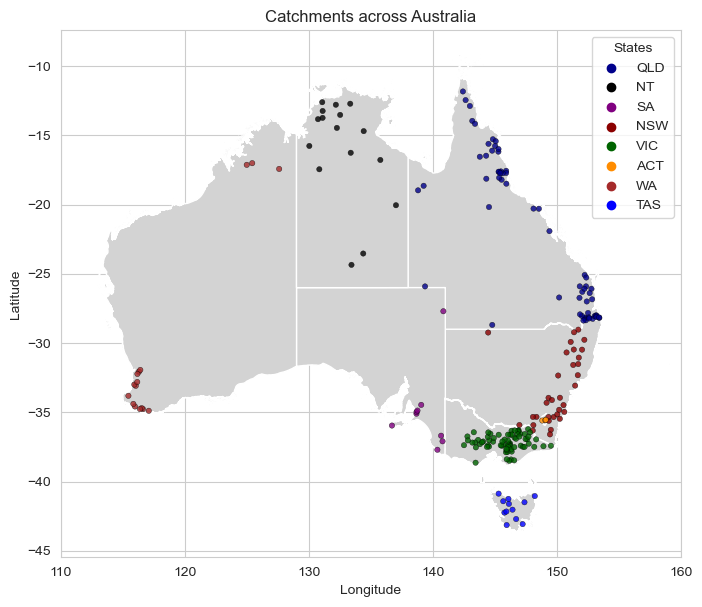}
\color{black}
\caption{Location of the Camels catchments across Australia across the different states.}
\label{fig:catchments}
\end{figure}

We use the  CAMELS dataset \cite{fowler2021camels} to demonstrate the use of deep learning models in streamflow prediction.  CAMELS   data contains hydrometeorological time series and landscape attributes for 222 catchments in Australia, Figure \ref{fig:catchments} presents an overview of the approximate locations.

 \begin{table*}[htbp]
    \centering 
   
    \caption{\textcolor{black}{Time series and static (landscape attribute)  in the CAMELS dataset selected for our study.}}
    \label{tab:summaryCAMELS}
    
    \begin{tabularx}{\textwidth}{l l l  l}  \hline
        {\bf Type} & {\bf Category/Description}   & {\bf Units} \\  \hline \hline
        Time Series & Streamflow &     mL $d^{-1}$\\ 
        Time Series & Precipitation &    mm $d^{-1}$ \\ 
        Time Series & Actual and potential evapotranspiration  & mm $d^{-1}$ \\ 
        Time Series & Maximum Temperature &   °C \\ 
        Time Series & Minimum Temperature &   °C \\ 
        Static & Mean streamflow &   mm $d^{-1}$ \\ 
        Static & Sensitivity of streamflow to rainfall &   - \\ 
        Static & Ratio of streamflow to precipitation &   - \\ 
        Static& Frequency of high flow days &   d y$^{-1}$ \\ 
        Static & Average duration of high flow &   days \\ 
        Static & Frequency of low flow days &   d y$^{-1}$ \\ 
        Static& Frequency of days with zero flow &  d y$^{-1}$ \\\hline

   \end{tabularx}
\end{table*}


\textcolor{black}{The CAMELS dataset is suitable for “large-sample hydrology” \cite{addor2020LSH, gupta2014LSH}  featuring time series of streamflow and 18 climatic variables at daily time steps. The time series data was recorded from  January 1911 and the addition of more stations over time featured a new beginning date per station. For example, for Station A, streamflow data would range from 1 January 1950 to 1 January 2020, but for Station B streamflow data would range from 1 January 1980 to 1 January 2020. Therefore, since our analysis is across stations, we must make a snapshot of the timeline with data present, and hence we selected data from 1st January 1980 January to 1st January  2020.}

The landscape attribute gives data and descriptions of attributes related to geology, soil, topography, land cover, anthropogenic influence and hydroclimatology for all 222 catchments. This is considered static information, where each catchment features only one value for each attribute. The metadata describes the latitude, longitude, and boundaries of each catchment. We provide a summary of the different attribute types in Table \ref{tab:summaryCAMELS}.

The dataset contains a variety of data that may not be relevant to the problem of flood prediction, so we select only the relevant time series data and static variables. In all our models, the input time series includes i.) daily cumulative precipitation, ii.) daily minimum air temperature, iii.) daily maximum air temperature, iv.) the time of year as sine/cosine functions as shown in Table \ref{tab:summaryCAMELS}.  Furthermore, we selected certain catchment characteristics for the models that require static data.  These include i.) mean streamflow, ii.) sensitivity of streamflow to rainfall. iii.) ratio of streamflow to precipitation. iv.) frequency of high flow days. v.) average duration of high flow. vi.) frequency of low flow days. vii.) frequency of days with zero flow as given in Table \ref{tab:summaryCAMELS}. These catchment attributes include climatic and vegetation indices, as well as soil and topographical properties.


\subsection{Data Reconstruction}\label{Data_Recon}

\begin{figure*}
\centering
\includegraphics[width=0.90\textwidth]{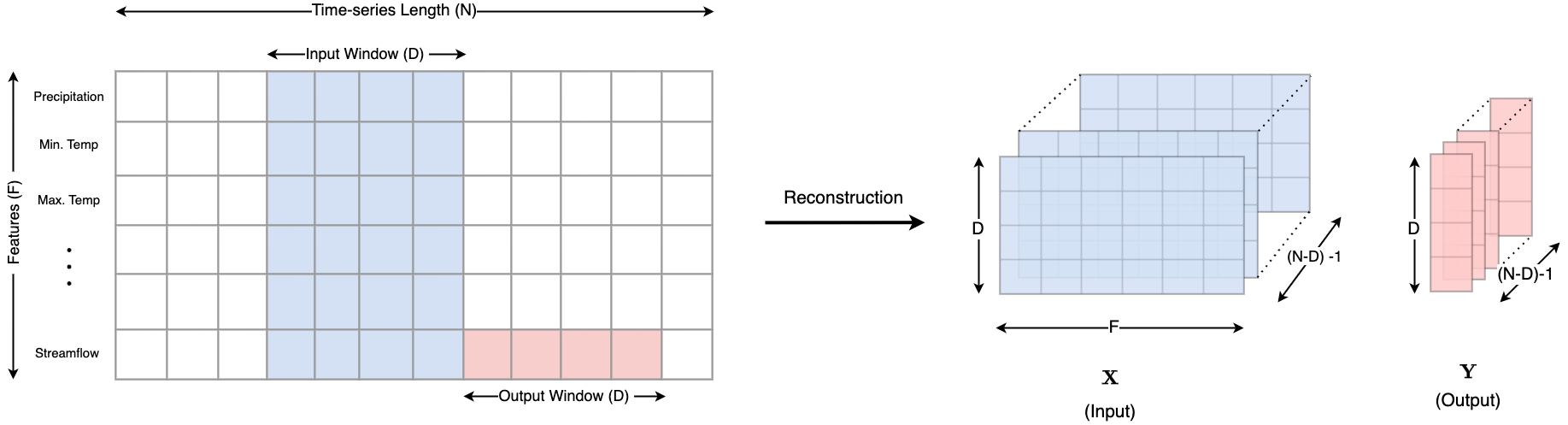}
\caption{Embedding: the transformation of a two-dimensional temporal dataset of features (daily time interval) into a three-dimensional format through a sliding window technique (windowing).}
\label{fig:embedding}
\end{figure*}

In order to use deep learning models, the original time-series data needs to be transformed (embedded) for both single-step and multistep-ahead prediction strategies. Therefore, the data needs to be processed using a sliding window approach, and the size of the window is crucial for this step. Taken's theorem demonstrates that any smooth coordinate transformation does preserve the important properties of the original time-series data \cite{takens1981detecting} with an appropriate embedding dimension. Thus, if the original univariate time-series data is of the form:  
\begin{equation}
  Y = [x_1, x_2, ..., x_t, ..., x_N]
\end{equation}

where, we observe $x_t$ at time $t$. We can create an \textit{embedded phase space} via sliding windows:
\begin{equation}
Y(t) = [x(t), x(t-T), x(t-2T), ..., x(t-(D-1)T)]
\end{equation}

where, $T$ is the time delay, $D$ is the embedding dimension (window-size); this is valid for all values of $t$ from $t = 0, 1, 2, ..., N- DT- 1$ and  $N$ is the length of the data. Assuming the original attractor is of dimension $d$, then $D = 2d + 1$ would be sufficient to capture the important characteristics of the original data in the embedded phase space \cite{takens1981detecting}. We use the embedded phase space to train the deep learning models, where $D$ represents the models' input size (neurons). Essentially, a univariate time series vector gets transformed into a matrix that is used for training and testing the model. Rather than using the entire univariate time series, the embedded phase space enables a sliding window approach for training the respective deep learning model, where the size of the sliding window defines the input. In the case of a multivariate time series, the same approach can be used, as shown in Figure \ref{fig:embedding}.

We need to transform the two-dimensional dataset of features and temporal sequences (daily time interval) into a three-dimensional format through a sliding window technique, as shown in Figure \ref{fig:embedding}. We use a sliding window   (size of $D \times F$) sliding over the 2-dimensional dataset given by  $F \times D$; where $F$ is the number of multivariate features. $N$ is the total length of the dataset, given by the number of days.



\begin{figure*}
\centering
\includegraphics[width=0.98\textwidth]{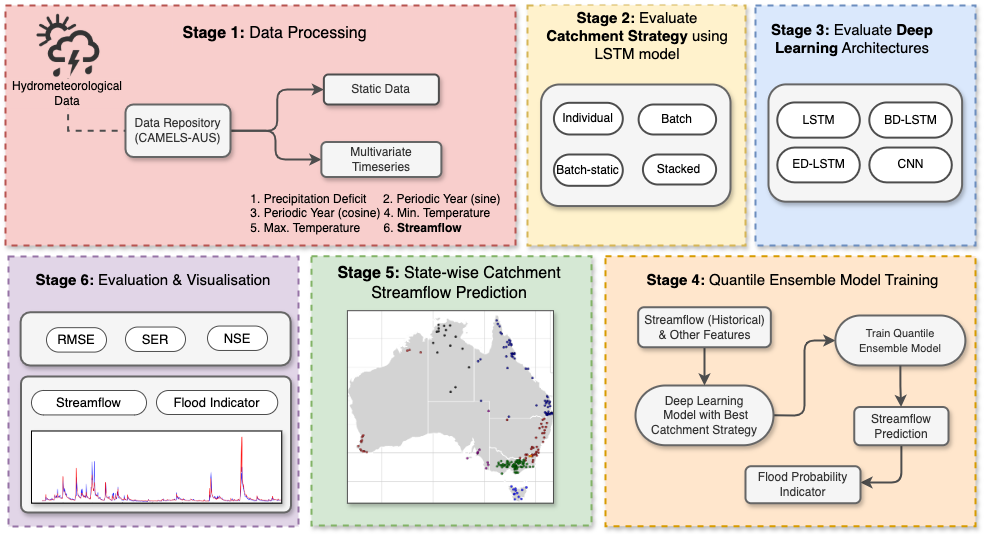}
\caption{Framework for comparing the different models for the different architectures (Stage 2 and 3) after acquiring and processing data (Stage 1). \color{black}{Stage 4 features our  ensemble quantile  deep learning model that predicts 2 extreme quantiles ($5^{th}$ and $95^{th}$ quantiles) along with regular streamflow prediction. 
We then utilise the model from Stage 4  to selected catchments across Australian states in Stage 5 and show results using the respective metrics (RMSE, NSE and SER). Finally, in Stage 6, we provide a hydrograph visualisation of flood probability and streamflow prediction for selected catchments.}}
\label{fig:framework1}
\end{figure*}


\begin{figure*}[htb]
    \centering
    \begin{subfigure}[b]{0.49\textwidth}
        \includegraphics[width=0.97\linewidth]{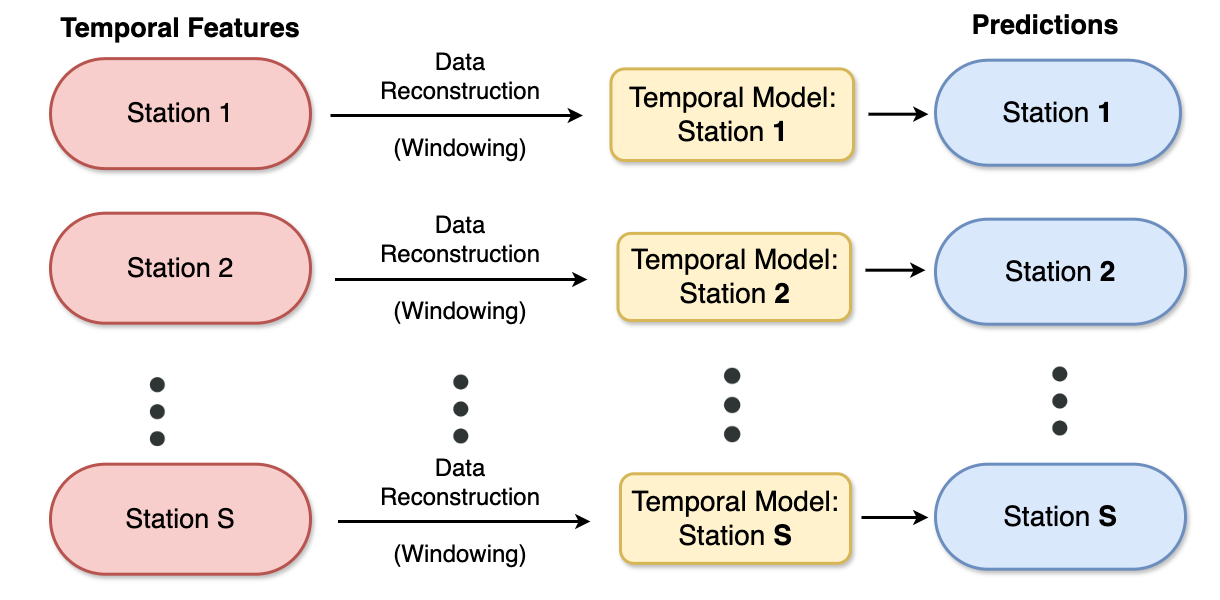}
        \caption{Individual models trained for each (individual) station separately.}
        \label{fig:individual}
    \end{subfigure}
    \begin{subfigure}[b]{0.49\textwidth}
        \includegraphics[width=0.97\linewidth]{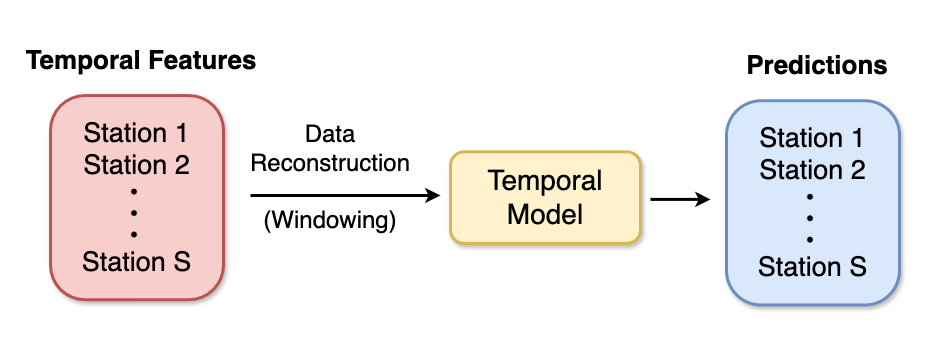}
        \caption{Batch-temporal: Single model for a subset of stations using temporal features only.}
        \label{fig:batch}
    \end{subfigure} 
     
    \begin{subfigure}[b]{0.48\textwidth}
        \includegraphics[width=0.97\linewidth]{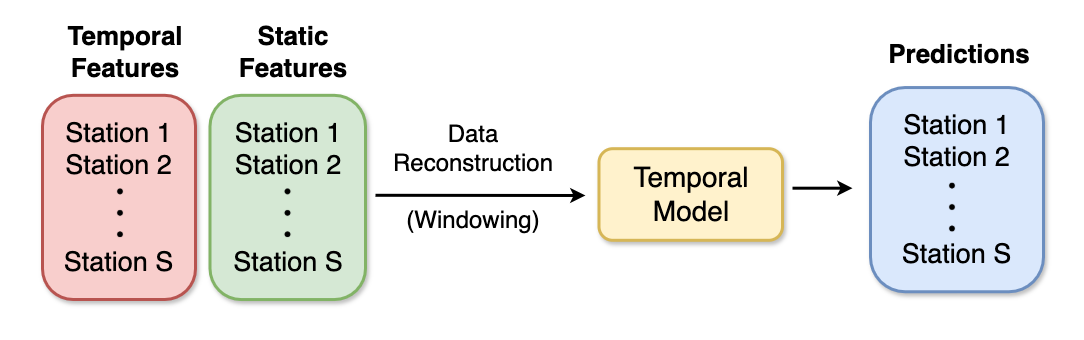}
        \caption{Batch-static: a single model for a subset of stations using both temporal and static features.}
        \label{fig:batch-static}
    \end{subfigure}
    \begin{subfigure}[b]{0.48\textwidth}
        \includegraphics[width=0.97\linewidth]{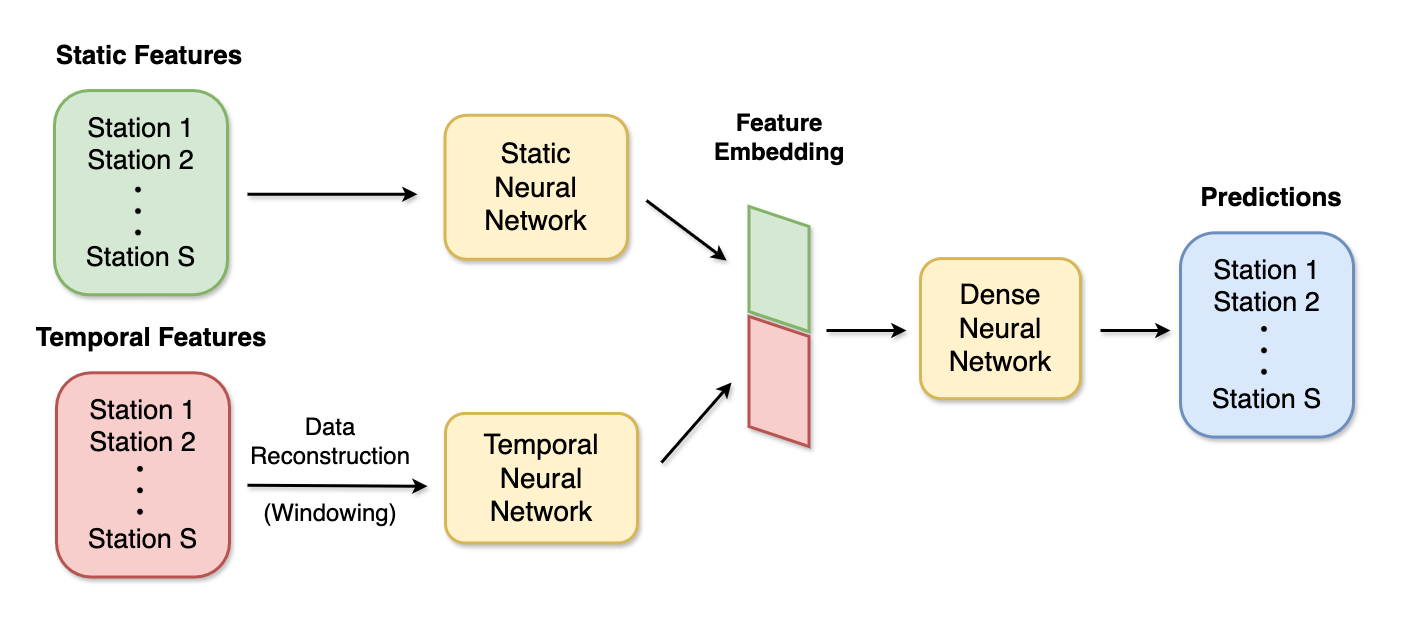}
        \caption{\color{black} Stacked Ensemble strategy:  Combining Temporal and Static models using a dense neural network.}
        \label{fig:batch-ensemble}
    \end{subfigure}
    \caption{Configurations}
    \label{fig:catchment-config}
\end{figure*}

\subsection{Framework}\label{Framework}

The contribution of our study is through fulfilling the following goals in order to accurately reproduce streamflow estimates for the given catchments.  

\begin{enumerate}[nosep]

  \item \textcolor{black}{Incorporate landscape attributes about the catchment (static input) as an additional feature with the time series data for the respective deep learning and ensemble learning models.}
  
  \item \textcolor{black}{Evaluate selected deep learning model and data configuration for streamflow prediction across selected Australian catchments.} 


  \item \textcolor{black}{Develop ensemble quantile-based deep learning model  for uncertainty quantification in streamflow prediction.}

  \item \textcolor{black}{Utilise streamflow prediction to create flood probability based on flood frequency analysis.}

  \item \textcolor{black}{Compare flood probability based on the historical flooding events of selected catchments for risk analysis.} 
\end{enumerate}

Figure \ref{fig:framework1} presents  the framework, which features   deep learning models with \textit{catchment data strategies} as given below: 

\begin{enumerate}

   \item The Individual strategy features a separate model for each base station, as shown in Figure \ref{fig:individual}.
    \item The Batch-Temporal strategy includes base stations from the entire region, where data from all the basins are concatenated for a single model as shown in Figure \ref{fig:batch}. 
    \item Batch-Static strategy features data from the entire region with static input, where each static input features landscape attributes as shown in Figure \ref{fig:batch-static}.  
    \item  The Ensemble catchment strategy (Figure \ref{fig:batch-ensemble}) combines forecasts from two models: one capturing temporal sequences and the other static features. These forecasts are concatenated and processed through a regression layer to generate final predictions. We investigate if the overall prediction accuracy improves by leveraging the strengths of both temporal and static feature models, enhancing the model's ability to make more accurate and robust predictions.

\end{enumerate}

 We begin with data processing that includes time series data reconstruction for use in deep learning models, as shown in Stage 1 of Figure \ref{fig:framework1}.
 We implement data reconstruction (windowing) and process precipitation-deficit, temperature, and antecedent streamflow as temporal data. The static data includes catchment streamflow properties such as frequency and duration of high/low streamflow and rainfall sensitivity.  In Stage 2, we assess the performance of four distinct catchment strategies: i.e. an Individual architecture, Batch-Temporal, Batch-Static, and Stacked-Ensemble model using an LSTM model. In Stage 3,  we evaluate the performance of four deep learning models including LSTM, CNN, bidirectional LSTM (BD-LSTM), and encoder-decoder LSTM (ED-LSTM). 
 
 \textcolor{black}{In Stage 4, we train the ensemble model, where we implement quantile regression in the output layer of the deep learning model (Figure \ref{fig:ensemble-proposed-architecture}). The ensemble features two Quantile-LSTM and a regular LSTM model to quantify the uncertainty in data via a confidence interval over the prediction horizon where the upper and lower bound is predicted by the Quantile-LSTM models, along with the conventional LSTM model for regular prediction.} 

 
 In Stage 5, we compare the respective models and apply the selected catchments
across the Australian states including NSW, Queensland, and Victoria. In Stage 5, we show results using the respective metrics. Finally, in Stage 6, we provide a hydrograph visualisation of flood probability and streamflow prediction for selected catchments.

\subsection{Quantile-LSTM: ensemble deep learning model}

Our ensemble deep learning model capitalises on quantile regression using an LSTM model known as Quantile-LSTM  (Figure \ref{fig:ensemble-proposed-architecture}). The Quantile-LSTM is based on the quantiles \cite{jia2022deep} of a distribution, which can enable a comprehensive understanding of prediction uncertainty when compared to the conventional LSTM model. We strategically use two separate Quantile-LSTM models to optimise for upper quantile $(q=0.95)$ and lower quantile $(q=0.05)$ of the  $90\%$ confidence interval. The confidence interval quantifies the data uncertainty in the streamflow prediction over the n-step prediction horizon. In addition, we train a regular-LSTM model using the mean-squared regression for predicting the expected value of streamflow over the prediction horizon. In order to identify flooding events, we first compute the flow-duration exceedance probability ($\alpha$), which is the cumulative frequency curve that shows the probability of streamflow exceeding a given threshold for the data period \cite{searcy1959flow}. In essence, this initial model predicts the probability that a flood of a certain magnitude occurs in the prediction window. We compute the annual flood exceedance probability using the historical streamflow data as:

\begin{eqnarray}
    \alpha = m / (n + 1)
\end{eqnarray}

 where, $m$ is the ranking from highest to lowest of annual maximum flow from the observed data, and $n$ is the total number of annual maximum streamflow cases. We identify the annual maximum flow with exceedance probability $20\%$, which is used as the \textit{flood threshold} ($\gamma$). Therefore, any event with streamflow higher than the \textit{flood threshold} is considered a flooding event.

\subsubsection{Flood prediction}

\begin{figure*}
\centering
\includegraphics[width=0.993\linewidth]{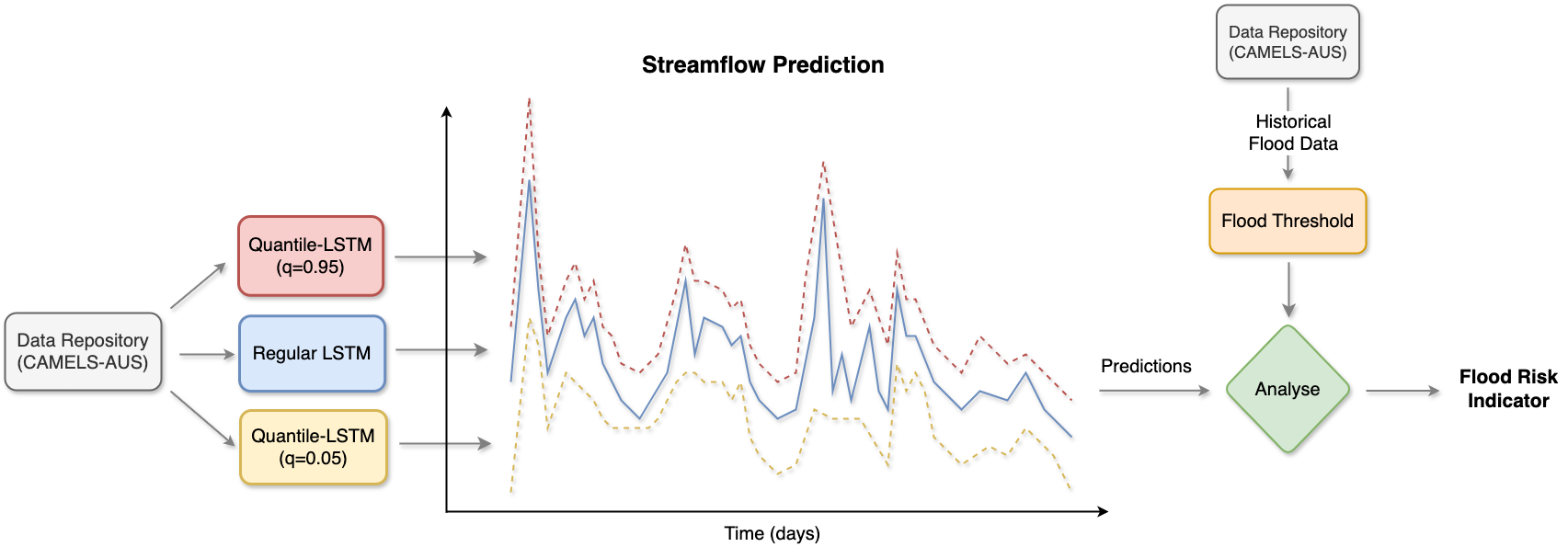}
\caption{Ensemble model utilising Quantile-LSTM for prediction of flood probability and streamflow with uncertainty bounds using quantile regression.} 
\label{fig:ensemble-proposed-architecture}
\end{figure*}

\textcolor{black}{ The flood risk indicator is a categorical value that classifies the likelihood of flooding based on the predicted streamflow values over the prediction horizon (n-steps) as shown in Figure \ref{fig:ensemble-proposed-architecture}. It is computed by comparing the predicted quantiles of streamflow (i.e. the expected value, $5^{th}$ percentile, and $95^{th}$  percentile predicted using the LSTM, Quantile-LSTM$_{q=.05}$, Quantile-LSTM$_{q=.95}$ models, respectively) against the predefined $\gamma$. For each of these models, we only consider the maximum predicted value over the prediction horizon to compute the \textit{flood risk indicator}. The classification follows a three-tier system: (1) 'High Risk' if the lower bound of the 90\% confidence interval exceeds the threshold ($Q_{q=0.05} > \gamma$), (2) 'Moderate Risk' if the expected value of streamflow surpasses the threshold ($Q_{reg} > \gamma$), and (3) 'Low Chance' if only the upper bound ($Q_{q=0.05} > \gamma$) exceeds the threshold. If none of these conditions are met, the chance of flooding would be deemed 'unlikely'.}

\subsection{Experiment Setup }\label{Framework-1}

\textcolor{black}{In this section, we provide technical details for implementing the conventional deep learning models (LSTM, BD-LSTM, ED-LSTM, and CNN). Moreover, we also provide the details for our proposed methods, which include the \textit{flood risk indicator} utilising Ensemble Quantile-LSTM.} We determined model hyperparameters such as hidden neurons, learning rate and epochs from trial experiments, as shown in Table \ref{tab:model-configuration}. We note that all the tested models implement a multi-step time series prediction with 5 inputs and 5 outputs (time steps). We train the model with a maximum time \textcolor{black}{of 150 epochs} using an Adam optimiser \cite{kingma2014adam} with default hyperparameters, such as the initial learning rate. 

 We scaled the time-series data using the min-max-scaler, i.e. scaled in the range [0,1]. In each basin, there are about 10,000 data points representing daily data spanning from 1990 to 2020.  We used the first 60 \% of the data from 1990 as  training data and the remaining for testing. We note that although there may be a climate bias in taking a block of data spanning more than a decade, this approach ensures reproducibility in results when compared to shuffling the data.   

It is necessary to use a variety of performance metrics for model benchmarking, as no one evaluation metric can fully capture the consistency, reliability, accuracy, and precision of a stream flow model.  The Nash–Sutcliffe-Efficiency (NSE) is the hydrologic analogue of the coefficient of determination ($R^2$ score), which can be less than zero. Silva et al. \cite{silva2022model} have suggested the Squared Error Relevance  (SER) - (SER-$\alpha$\%) for extreme events. The SER calculates the Root Mean Squared Error (RMSE)  of test windows that contain the $\alpha$ percentile of the data, measuring the ability of a model to predict rare data points. For example, SER-1\% would find the RMSE of windows that contain the 1\% most extreme stream flows for a given basin. Therefore, we use the following metrics to evaluate prediction performance accuracy over multiple steps (prediction horizon).

\begin{enumerate}[nosep]

   \item  The RMSE is  given by  $\sqrt{\sum_{i=1}^{N}(\hat{y_i}-y_i)^2}$  between predicted $\hat{y_i}$ and actual $y_i$ values for $N$ data points. 
 
    \item The  SER is  given by  $i\%$ RMSE   $\sqrt{\sum_{i\in D^i}(\hat{y_i}-y_i)^2}$  of the windows that contain stream flows above the i\textit{th} percentile in the test dataset, $i\in(1,2,5,10,25,50,75)$.

   \item The  NSE is given by  $1 - \frac{\sum_{i=1}^{N}(Q^t_o-Q^t_m)^2}{\sum_{i=1}^{N}(Q^t_o-\bar{Q^t_o})^2}$   and ranges from (-$\infty$, 1).
\end{enumerate}

\begin{table*}[htbp!] 
    \centering
    \caption{Configuration for conventional deep learning and conditional ensemble model.}
    \label{tab:model-configuration}
    \begin{tabularx}{\textwidth}{l l l}  \hline
         Models &  Hidden Units & Type  \\\hline
        LSTM &  20  & Conventional Model\\ 
        BD-LSTM & 20  &  Conventional Model \\ 
        ED-LSTM & 20 &  Conventional Model\\  
        CNN & 64 &  Conventional Model \\ \hline  
          
        Quantile-LSTM$_{q=0.05}$ & 20 cells   &  Quantile Ensemble\\ 
        Quantile-LSTM$_{q=0.95}$ & 20 cells   & Quantile ensemble\\  
        LSTM & 20 cells   &  Quantile Ensemble\\ \hline
    \end{tabularx}
\end{table*}

 \begin{table*}[htbp!]
    \centering
    \caption{Accuracy (SER) on the test data of catchment strategies using LSTM model  for South Australia}
    \label{tab:catchment}
    \begin{tabularx}{\textwidth}{c c c c c c c c c}  \hline
        {\bf Architecture} & {\bf SER 1\%} & {\bf SER 2\%} & {\bf SER 5\%} & {\bf SER 10\%} & {\bf SER 25\%} & {\bf SER 50\%} & {\bf SER 75\%} & {\bf RMSE}\\\hline
        Individual & 1.8321 & 1.1904 & 0.6606 & 0.4165 & 0.2226 & 0.1337 & 0.1064 & 0.0963 \\
         & ±0.0412 &  ±0.0221 & ±0.0143 & ±0.0117 & ±0.0082 & ±0.0072 & ±0.0069 & ±0.0068 \\\hline
        Batch-Temporal & 2.2775 & 1.6474 & 1.0696 & 0.6884 & 0.3669
 & 0.2332 & 0.1915 & 0.1789\\
         & ±0.1107 &  ±0.0884 & ±0.1174 & ±0.0670 & ±0.0343 & ±0.0229
 & ±0.0203 & ±0.0200 \\\hline
        Batch-Static & 1.9712 & 1.2450 & 0.6755 & 0.4853 & 0.2312
	& 0.1461 & 0.1127 & 0.1040\\ 
          & ±0.2421 &  ±0.1556 & ±0.0805 & ±0.0459 & ±0.0227 & ±0.0164 & ±0.0134 & ±0.0121 \\\hline
        Stacked Ensemble &  4.1612 & 2.6328 & 1.3091 & 0.7229 & 0.3363
 & 0.1975 & 	0.1627 & 0.1499\\ 
         & ±0.0445 &  ±0.0322 & ±0.0181 & ±0.0109 & ±0.0050 & ±0.0024 & ±0.0017 & ±0.0014 \\\hline
    \end{tabularx}
\end{table*}

  \begin{table*}[!ht]
    \centering 
    \caption{Architecture wise- model comparisons showing test accuracy (SER) to track best-performing model (South Australia)}
    \label{tab:dlmodels}
    \begin{tabularx}{\textwidth}{c c c c c c c c c}  \hline
        {\bf Model} & {\bf SER 1\%} & {\bf SER 2\%} & {\bf SER 5\%} & {\bf SER 10\%} & {\bf SER 25\%} & {\bf SER 50\%} & {\bf SER 75\%} & {\bf RMSE}\\\hline
      LSTM & 1.8321 & 1.1904 & 0.6606 & 0.4165 & 0.2226 & 0.1337 & 0.1064 & 0.0963 \\
         & ±0.0412 &  ±0.0221 & ±0.0143 & ±0.0117 & ±0.0082 & ±0.0072 & ±0.0069 & ±0.0068 \\\hline
     BD-LSTM & 1.7041 & 1.1264 & 0.6426 & 0.4029 & 0.2178 & 0.1384 & 0.1143 & 0.1053\\
         & ±0.0529 &  ±0.0383 & ±0.0309 & ±0.0222 & ±0.0123 & ±0.0092 & ±0.0089 & ±0.0089 \\\hline
      CNN & 1.7349 & 1.1204 & 0.6199 & 0.3776 & 0.1964 & 0.1273 & 0.1056 & 0.0976\\
         & ±0.0564 &  ±0.0288 & ±0.0158 & ±0.0085 & ±0.0035 & ±0.0018 & ±0.0016 & ±0.0016 \\\hline
        ED-LSTM & 1.7468 & 1.1528 & 0.6508 & 0.4071 & 0.2188 & 0.1390 & 0.1144 & 0.1055\\
         & ±0.0251 &  ±0.0183 & ±0.0199 & ±0.0168 & ±0.0121 & ±0.0100 & ±0.0094 & ±0.0092 \\\hline
     Quantile-LSTM$_{q=0.05}$ & 2.0646 & 1.2959 & 0.6683 & 0.3790 & 0.1860 & 0.1109 & 0.0884 & 0.0801 \\
                    & ±0.0580 & ±0.0306 & ±0.0154 & ±0.0091 & ±0.0045 & ±0.0031 & ±0.0028 & ±0.0027 \\ \hline
     Quantile-LSTM$_{q=0.95}$ & 1.5932 & 1.0813 & 0.6585 & 0.4390 & 0.2443 & 0.1478 & 0.1166 & 0.1051 \\
                    & ±0.0459 & ±0.0306 & ±0.0169 & ±0.0180 & ±0.0114 & ±0.0083 & ±0.0080 & ±0.0080 \\ \hline
    \end{tabularx}
\end{table*}

\begin{table*}[htb]
    \centering
    \caption{Ensemble Quantile-LSTM model results showing test accuracy (SER)  for major states in Australia}
    \label{tab:allstates}
    \begin{tabular}{llrrrrrrrr}
    \toprule
     {\bf State} & {\bf Ensemble Quantile LSTM} & {\bf SER 1\%} & {\bf SER 2\%} & {\bf SER 5\%} & {\bf SER 10\%} & {\bf SER 25\%} & {\bf SER 50\%} & {\bf SER 75\%} & {\bf RMSE} \\
    \midrule
    \multirow[t]{6}{*}{NT} & Regular LSTM & 1.2810 & 0.8192 & 0.4845 & 0.3194 & 0.1621 & 0.0827 & 0.0618 & 0.0519 \\
                           & & ±0.0262 & ±0.0190 & ±0.0124 & ±0.0078 & ±0.0032 & ±0.0016 & ±0.0011 & ±0.0008 \\
                           \cline{2-10}
                           &Quantile LSTM q=0.05 & 1.8632 & 1.2321 & 0.7665 & 0.5174 & 0.2520 & 0.1275 & 0.0963 & 0.0816 \\
                           &  & ±0.0078 & ±0.0057 & ±0.0041 & ±0.0030 & ±0.0016 & ±0.0008 & ±0.0005 & ±0.0004 \\
                           \cline{2-10}
                           &Quantile LSTM q=0.95 & 1.1136 & 0.7152 & 0.4377 & 0.3100 & 0.1810 & 0.0946 & 0.0705 & 0.0590 \\
                           &  & ±0.0156 & ±0.0097 & ±0.0074 & ±0.0092 & ±0.0093 & ±0.0053 & ±0.0036 & ±0.0028 \\
    \cline{1-10}
    \multirow[t]{6}{*}{NSW} & Regular LSTM & 1.3857 & 0.9052 & 0.4724 & 0.2709 & 0.1214 & 0.0654 & 0.0449 & 0.0352 \\
                            & & ±0.0138 & ±0.0100 & ±0.0070 & ±0.0050 & ±0.0021 & ±0.0010 & ±0.0008 & ±0.0006 \\
                            \cline{2-10}
                            &Quantile LSTM q=0.05 & 1.5695 & 1.0295 & 0.5494 & 0.3163 & 0.1412 & 0.0752 & 0.0512 & 0.0398 \\
                            & & ±0.0144 & ±0.0116 & ±0.0085 & ±0.0050 & ±0.0020 & ±0.0010 & ±0.0007 & ±0.0005 \\
                            \cline{2-10}
                            &Quantile LSTM q=0.95 & 1.2753 & 0.8306 & 0.4350 & 0.2557 & 0.1194 & 0.0655 & 0.0452 & 0.0355 \\
                            & & ±0.0126 & ±0.0085 & ±0.0058 & ±0.0054 & ±0.0036 & ±0.0020 & ±0.0014 & ±0.0011 \\
    \cline{1-10}
    \multirow[t]{6}{*}{ACT} & Regular LSTM & 1.7718 & 1.1026 & 0.4953 & 0.2621 & 0.1182 & 0.0623 & 0.0423 & 0.0325 \\
                            &  & ±0.0431 & ±0.0230 & ±0.0097 & ±0.0055 & ±0.0026 & ±0.0015 & ±0.0011 & ±0.0009 \\
                            \cline{2-10}
                            &Quantile LSTM q=0.05 & 1.9456 & 1.1943 & 0.5291 & 0.2756 & 0.1222 & 0.0637 & 0.0427 & 0.0326 \\
                            &  & ±0.0221 & ±0.0104 & ±0.0059 & ±0.0036 & ±0.0017 & ±0.0009 & ±0.0006 & ±0.0005 \\
                            \cline{2-10}
                            &Quantile LSTM q=0.95 & 1.5312 & 1.0087 & 0.4852 & 0.2759 & 0.1355 & 0.0756 & 0.0529 & 0.0412 \\
                            &  & ±0.0336 & ±0.0231 & ±0.0135 & ±0.0101 & ±0.0064 & ±0.0042 & ±0.0031 & ±0.0025 \\
    \cline{1-10}
    \multirow[t]{6}{*}{WA} & Regular LSTM & 2.2494 & 1.8241 & 1.1968 & 0.8022 & 0.4337 & 0.2719 & 0.2097 & 0.1809 \\
                           &  & ±0.0519 & ±0.0396 & ±0.0229 & ±0.0125 & ±0.0034 & ±0.0053 & ±0.0039 & ±0.0032 \\
                           \cline{2-10}
                           &Quantile LSTM q=0.05 & 3.3430 & 2.7612 & 1.9196 & 1.2671 & 0.6159 & 0.3499 & 0.2675 & 0.2316 \\
                           &  & ±0.0239 & ±0.0200 & ±0.0161 & ±0.0126 & ±0.0068 & ±0.0038 & ±0.0026 & ±0.0019 \\
                           \cline{2-10}
                           &Quantile LSTM q=0.95 & 1.7287 & 1.4083 & 0.9988 & 0.7465 & 0.5152 & 0.3706 & 0.2869 & 0.2423 \\
                           &  & ±0.0327 & ±0.0228 & ±0.0152 & ±0.0126 & ±0.0166 & ±0.0179 & ±0.0148 & ±0.0122 \\
    \cline{1-10}
    \multirow[t]{6}{*}{SA} & Regular LSTM & 1.5887 & 1.0652 & 0.5973 & 0.3534 & 0.1778 & 0.0930 & 0.0633 & 0.0528 \\
                            &  & ±0.0235 & ±0.0158 & ±0.0111 & ±0.0084 & ±0.0056 & ±0.0031 & ±0.0021 & ±0.0018 \\
                            \cline{2-10}
                           &Quantile LSTM q=0.05 & 2.2325 & 1.4956 & 0.8305 & 0.4787 & 0.2226 & 0.1151 & 0.0789 & 0.0674 \\
                           &  & ±0.0124 & ±0.0093 & ±0.0062 & ±0.0040 & ±0.0018 & ±0.0009 & ±0.0006 & ±0.0005 \\
                           \cline{2-10}
                           &Quantile LSTM q=0.95 & 1.4409 & 1.0195 & 0.6268 & 0.4112 & 0.2318 & 0.1249 & 0.0854 & 0.0717 \\
                           &  & ±0.0341 & ±0.0265 & ±0.0182 & ±0.0144 & ±0.0137 & ±0.0081 & ±0.0054 & ±0.0043 \\
    \cline{1-10}
    \multirow[t]{6}{*}{QLD} & Regular LSTM& 1.2907 & 0.8690 & 0.5061 & 0.3282 & 0.1559 & 0.0827 & 0.0557 & 0.0422 \\
                            &  & ±0.0150 & ±0.0110 & ±0.0065 & ±0.0041 & ±0.0017 & ±0.0008 & ±0.0005 & ±0.0004 \\
                            \cline{2-10}
                            &Quantile LSTM q=0.05 & 1.6486 & 1.1355 & 0.6691 & 0.4291 & 0.2003 & 0.1053 & 0.0708 & 0.0536 \\
                            &  & ±0.0183 & ±0.0127 & ±0.0088 & ±0.0057 & ±0.0026 & ±0.0013 & ±0.0009 & ±0.0007 \\
                            \cline{2-10}
                            & Quantile LSTM q=0.95 & 1.0673 & 0.7300 & 0.4512 & 0.3187 & 0.1709 & 0.0947 & 0.0645 & 0.0491 \\
                            &  & ±0.0108 & ±0.0051 & ±0.0047 & ±0.0050 & ±0.0034 & ±0.0019 & ±0.0013 & ±0.0010 \\
    \cline{1-10}
    \bottomrule
    \end{tabular}
\end{table*}

\section{Results}\label{First-Objective-Results}



We first evaluate four distinct catchment modelling strategies using a conventional LSTM model to identify the optimal catchment modelling strategy using a subset of the data from South Australia as a case study. We compare the Individual, Batch-Temporal, Batch-Static and Ensemble catchment modelling strategies. The analysis considers metrics such as SER\% (with percentile cuts at 1, 2, 5, 10, 50, and 75) and RMSE mean across five prediction horizons. This enables a thorough evaluation of the influence of extreme flood conditions while making comparisons across different models and architectures. Table \ref{tab:catchment}  presents the results, highlighting that the Individual catchment strategy emerged as the most effective across various metrics (SER percentiles).  This strategy has the lowest (best) RMSE, which indicates that separate models for individual gauging stations provide the best performance.


The next step is to compare deep learning models for catchments in South Australia, including the  LSTM, BD-LSTM, CNN, ED-LSTM, and Ensemble Quantile-LSTM. We chose the South Australia dataset for comprehensive evaluation since it has a lower number of gauging stations and data when compared to large states such as New South Wales.  As shown in Table \ref{tab:dlmodels}, we find that the Quantile-LSTM (q=0.05) provides the best prediction accuracy (RMSE of 0.0801) and SER with high $\alpha$ values (50 and 75). This is close to the LSTM model  (RMSE of 0.0963), and CNN (RMSE of 0.0976), followed by BD-LSTM and ED-LSTM which give the lowest performance, respectively. The Quantile-LSTM (q=0.95) which aims to capture the upper quantile of the predictive interval performs the best near the peaks (SER 1\% and SER2\%). The results reinforce the efficacy of the Ensemble Quantile-LSTM model as a robust choice for hydrological modelling with uncertainty quantification in South Australia. This performance highlights the resilience and effectiveness in capturing extreme and regular streamflow trends, where extreme trends could be associated with flood occurrences.

Furthermore, we extend the Ensemble  Quantile-LSTM model across all states in Australia.  We only evaluate the performance of Ensemble Quantile-LSTM models on 5 selected basins from each state; i.e. we selected the stations with the highest runoff ratio in each state. Table \ref{tab:allstates} presents the results, highlighting that in some states, the prediction accuracy (RMSE) is better than others, which is correlated to the SER percentiles. The difference in the results across the states could be due to different levels of extreme values in the data, and also the size of the data and the presence of noise. The Northern Territory (NT) and Western Australia (WA) have the worst model accuracy (RMSE), while NSW, Australian Capital Territory (ACT), Queensland (QLD), and Victoria (VIC) have some of the best prediction accuracies (Table \ref{tab:allstates}). Notably, NSW and Queensland feature the highest number of catchments  (gauging stations) and are also the states that are prone to extreme floods \cite{gu2020changing}. 

\begin{figure*}[htb]
    \centering
    \begin{subfigure}[b]{0.98\textwidth}
        \centering
        \includegraphics[width=0.95\linewidth]{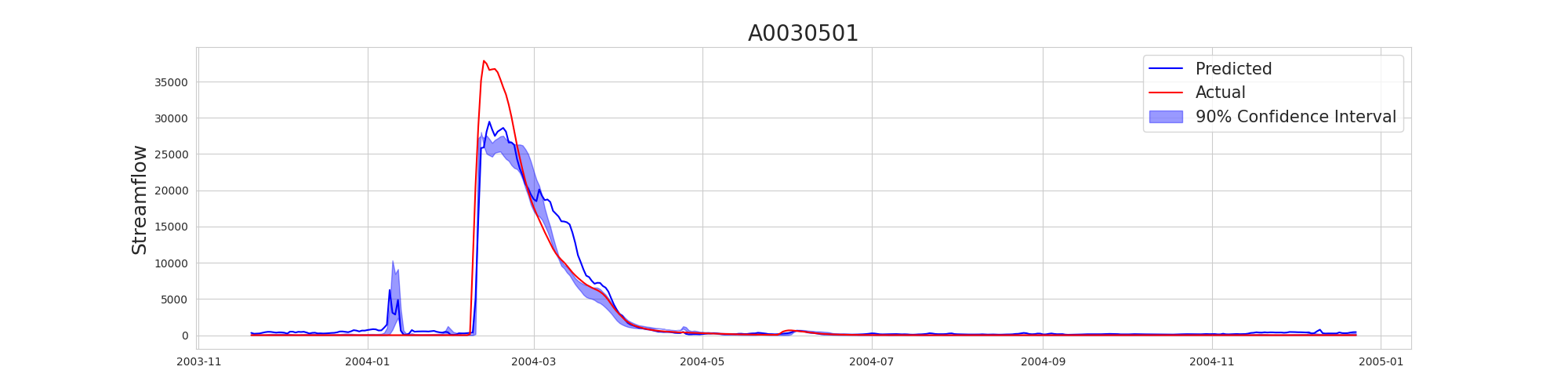}
        \caption{\textcolor{black}{South Australia Basin ID:} A0030501}
    \end{subfigure}
    
    \begin{subfigure}[b]{0.98\textwidth}
        \centering
        \includegraphics[width=0.95\linewidth]{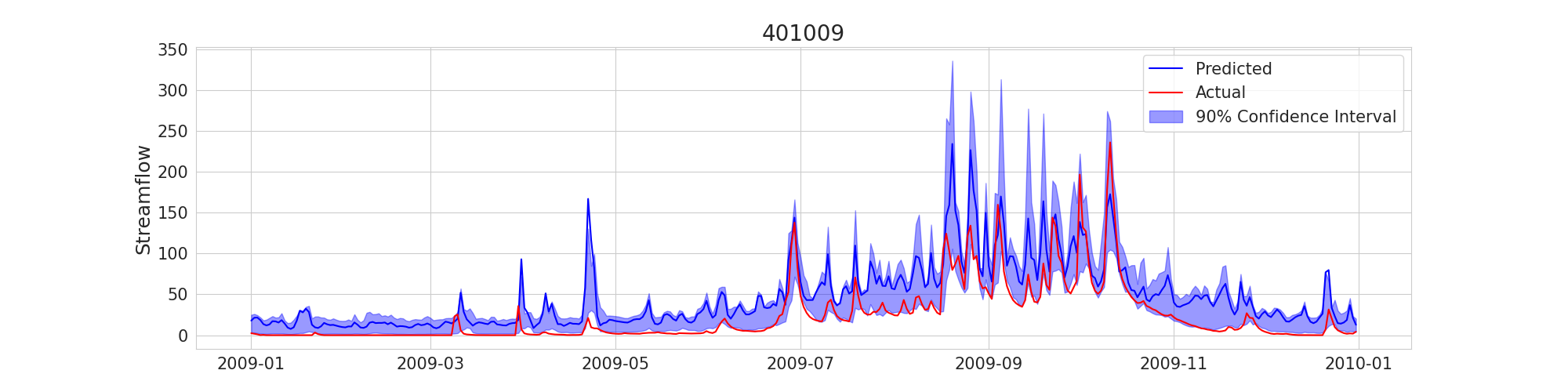}
        \caption{\textcolor{black}{New South Wales Basin ID:} 401009}
    \end{subfigure}
    
    \begin{subfigure}[b]{0.98\textwidth}
        \centering
        \includegraphics[width=0.95\linewidth]{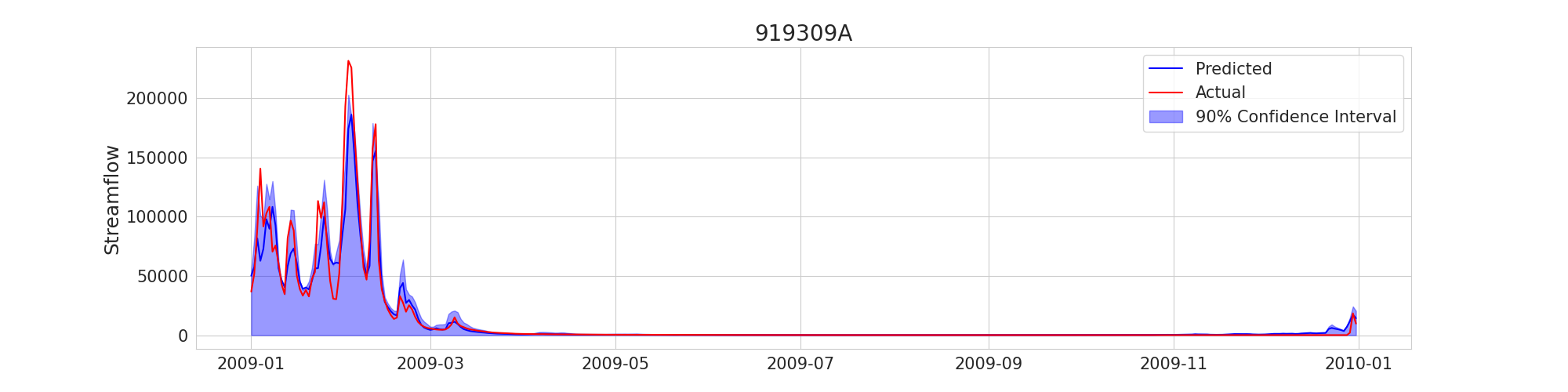}
        \caption{\textcolor{black}{Queensland Basin ID:} 919309A}
    \end{subfigure}

    \begin{subfigure}[b]{0.98\textwidth}
        \centering
        \includegraphics[width=0.95\linewidth]{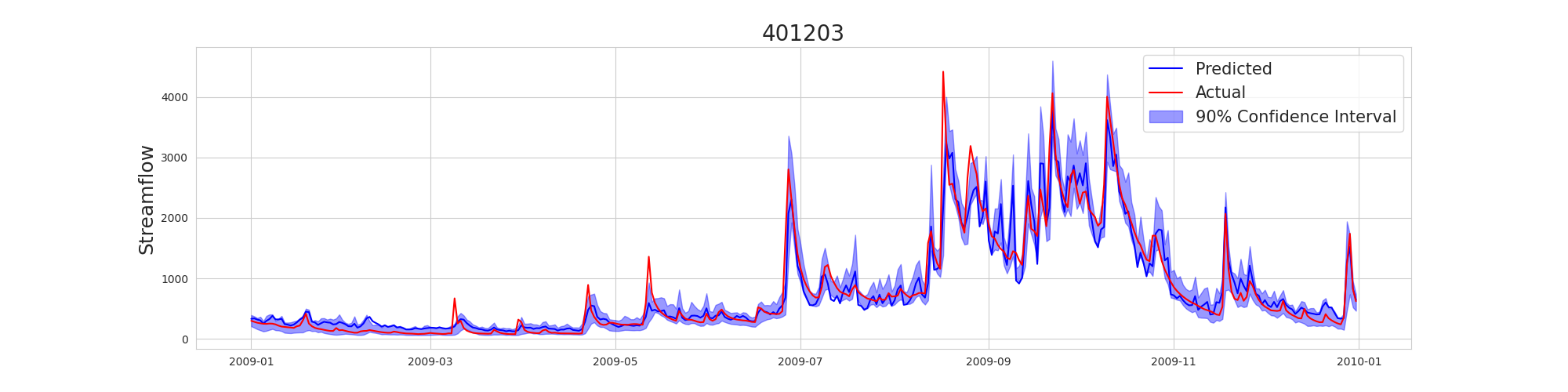}
        \caption{\textcolor{black}{Victoria Basin ID:} 401203}
    \end{subfigure}
    
    \caption{Ensemble Quantile-LSTM streamflow predictions for a randomly selected basins from South Australia, Queensland, New South Wales, and Victoria}
    \label{fig:predictions_1}
\end{figure*}

\begin{figure*}[htb]
    \centering
    \begin{subfigure}[b]{0.98\textwidth}
        \centering
        \includegraphics[width=0.95\linewidth]{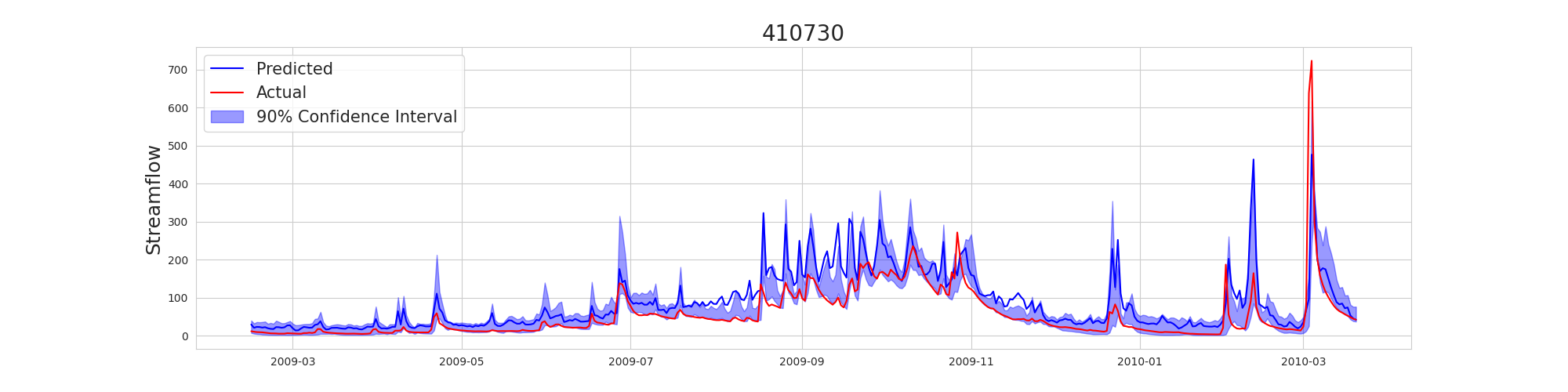}
        \caption{\textcolor{black}{Australian Capital Territory Basin ID:} 410730}
    \end{subfigure}
    
    \begin{subfigure}[b]{0.98\textwidth}
        \centering
        \includegraphics[width=0.95\linewidth]{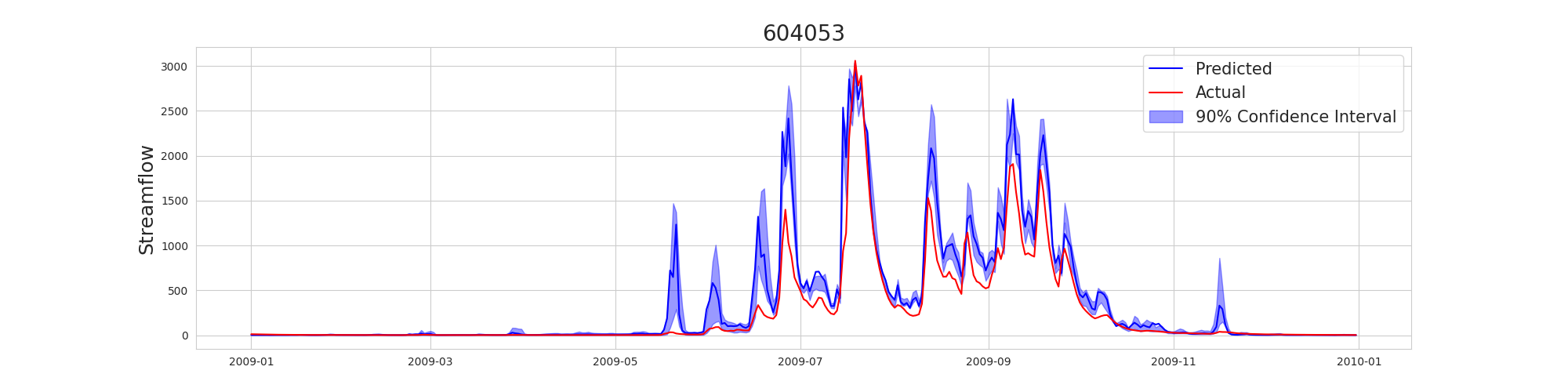}
        \caption{\textcolor{black}{Western Australia Basin ID:} 604053}
    \end{subfigure}
    
    \begin{subfigure}[b]{0.98\textwidth}
        \centering
        \includegraphics[width=0.95\linewidth]{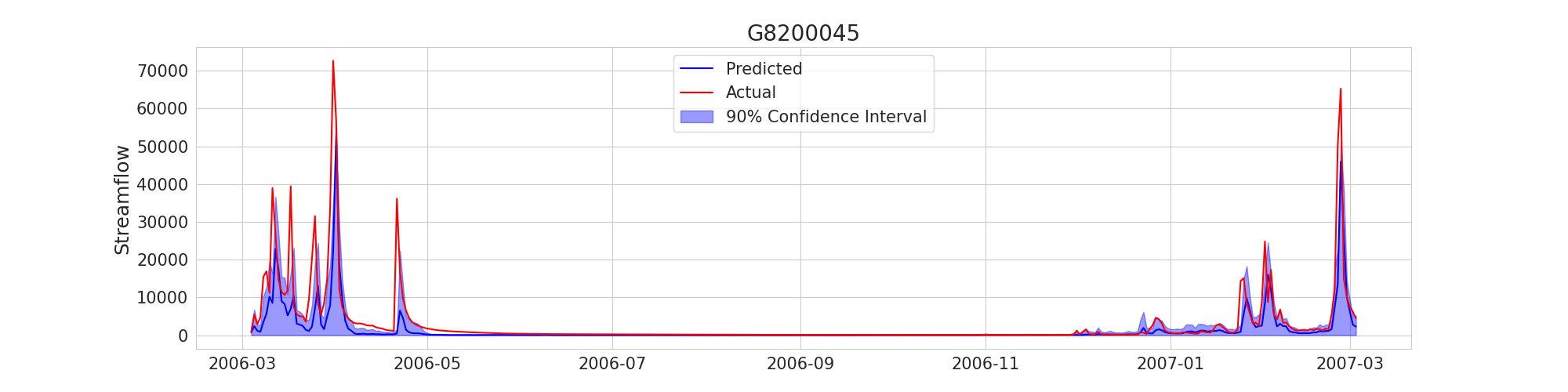}
        \caption{\textcolor{black}{Northern Territory Basin ID:} G8200045}
    \end{subfigure}

    \begin{subfigure}[b]{0.98\textwidth}
        \centering
        \includegraphics[width=0.95\linewidth]{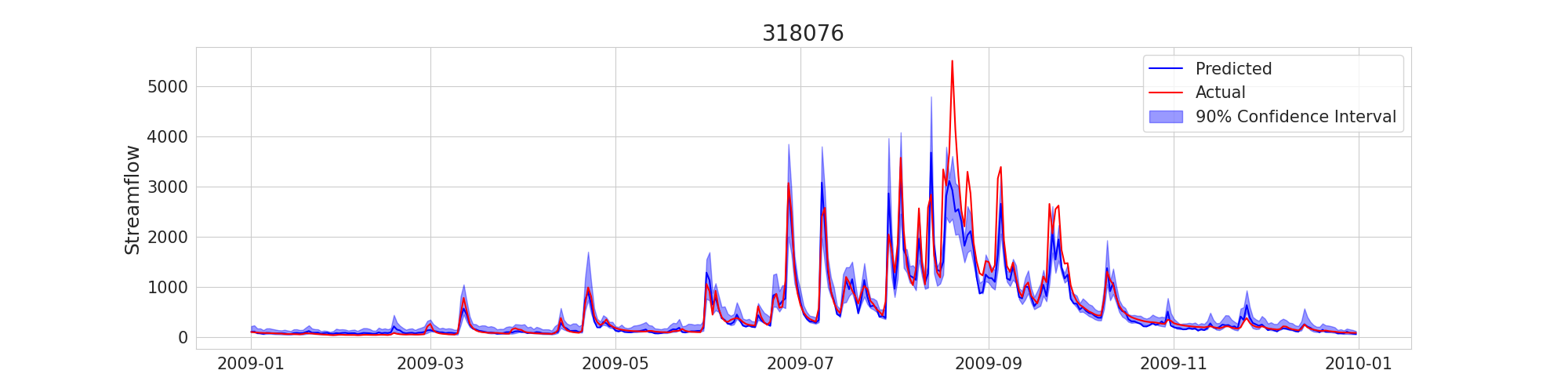}
        \caption{\textcolor{black}{Tasmania Basin ID:} 318076}
    \end{subfigure}
    
    \caption{Ensemble Quantile-LSTM streamflow predictions for randomly selected basins from Australian Capital Territory, Western Australia, Northern Territory, and Tasmania}
    \label{fig:predictions_2}
\end{figure*}

 In Figures \ref{fig:predictions_1} to  \ref{fig:predictions_2}, we provide a visualisation of actual and predicted values using the  Ensemble  Quantile-LSTM model for selected gauging stations (showing Basin ID) for the different states for one year. In most cases, we notice that short-duration extreme rainfall events that contribute to a high chance of flooding are well captured by the confidence interval predicted using the Ensemble Quantile-LSTM model. 
However, there are some cases where the actual streamflow crosses the 90\% confidence interval predicted by the Ensemble Quantile-LSTM model (e.g. Figure \ref{fig:predictions_1}(a) and \ref{fig:predictions_1}(c)). We also notice that in some basins where the flow is mostly zeros, the Quantile-LSTM (q=0.05) is biased towards zero flows (e.g. Figure \ref{fig:predictions_1}(c) and \ref{fig:predictions_2}(c)).

\begin{figure*}[htb]
    \centering
    \begin{subfigure}[b]{0.98\textwidth}
        \centering
        \includegraphics[width=0.95\linewidth]{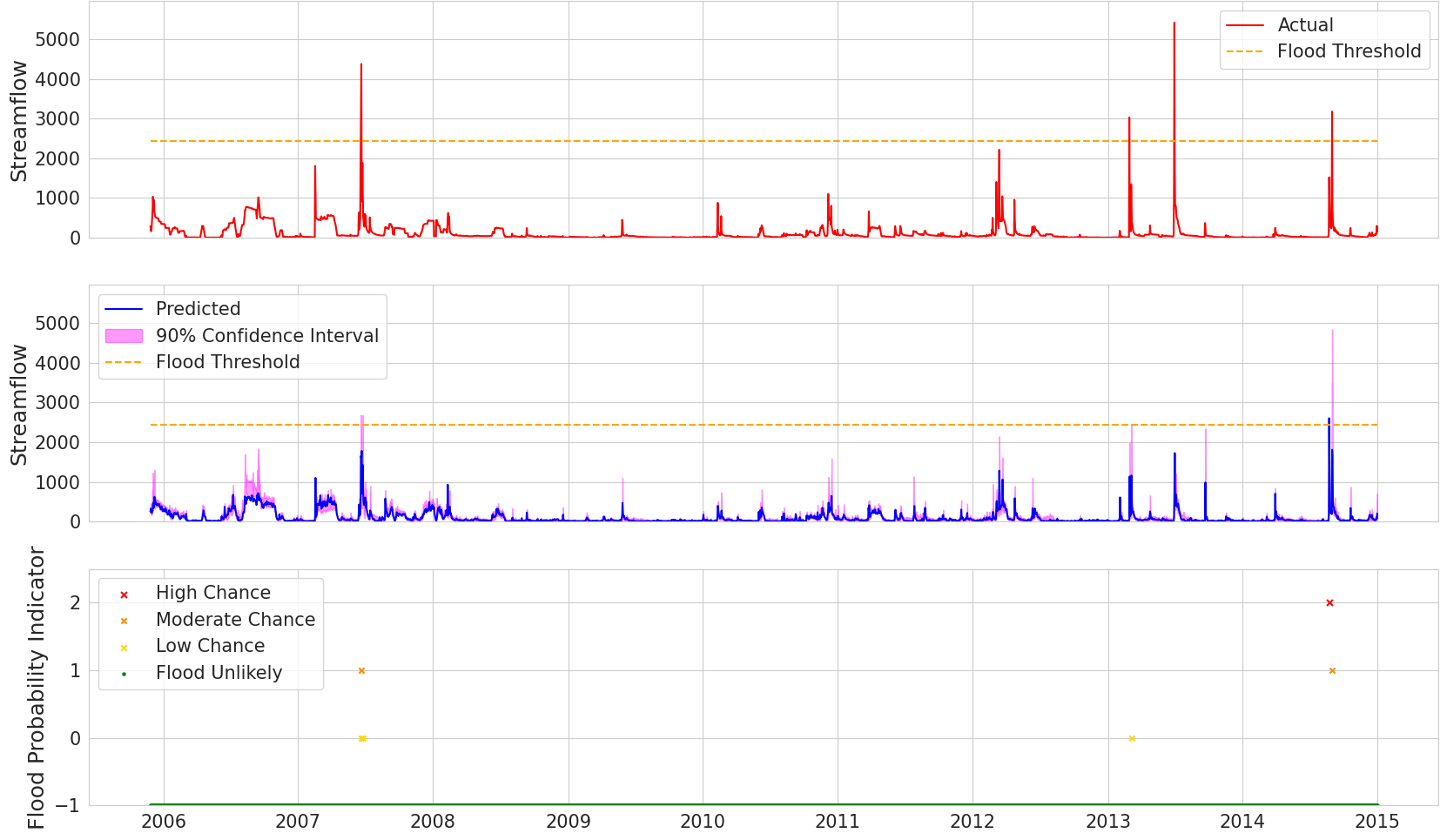}
        \caption{Nepean River at Maguires Crossing (Station ID: 212209) in Hawkesbury River region of New South Wales}
    \end{subfigure}
    \begin{subfigure}[b]{0.98\textwidth}
        \centering
        \includegraphics[width=0.95\linewidth]{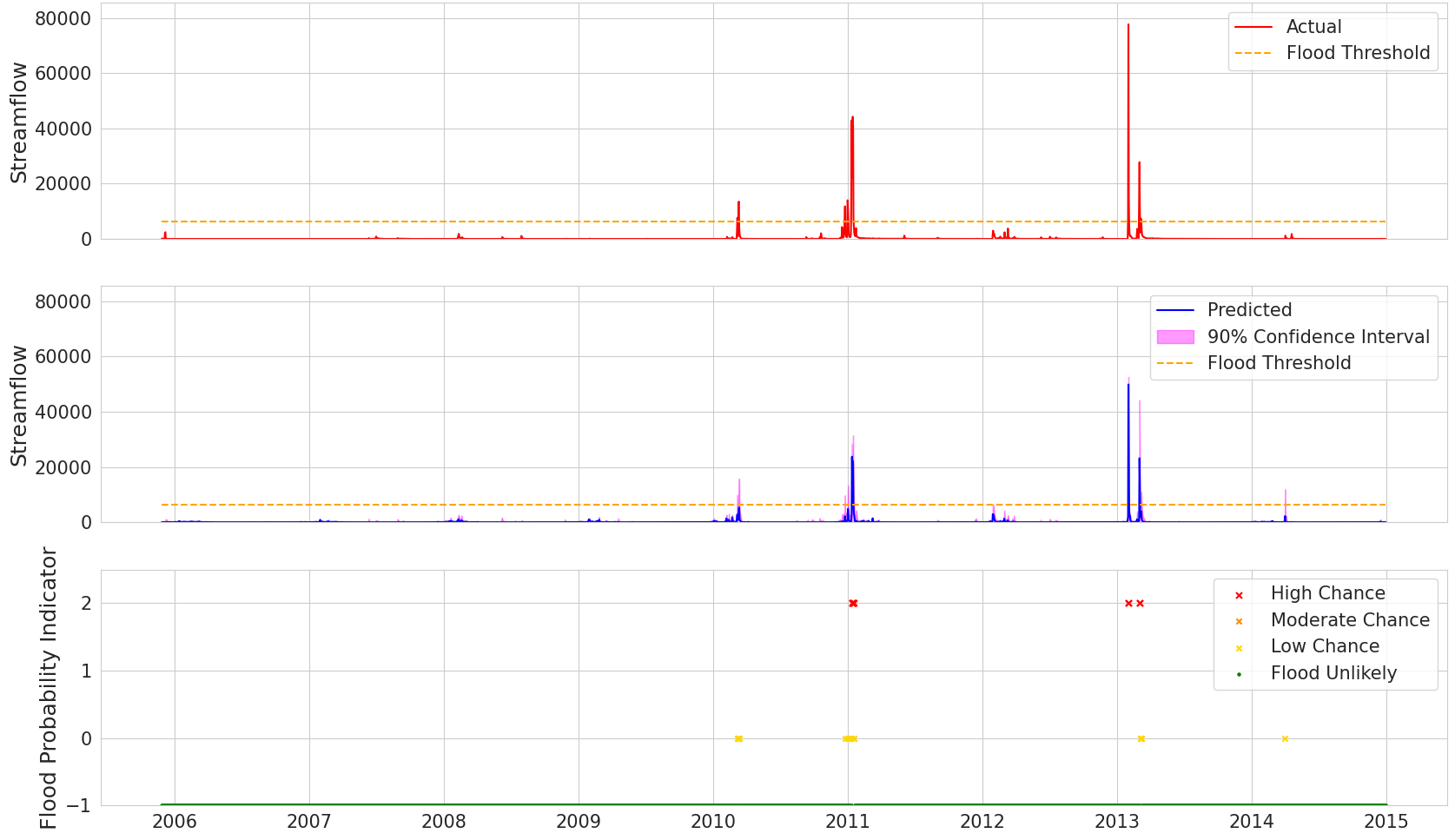}
        \caption{Barambah Creek at Litzows (Station ID: 136202D) in Burnett river region of Queensland}
    \end{subfigure}
    \caption{Flooding events in selected catchments of New South Wales and Queensland predicted using the Ensemble Quantile-LSTM streamflow prediction model }
    \label{fig:flood_pred_1}
\end{figure*}

\begin{figure*}[htb]
    \centering
    \begin{subfigure}[b]{0.98\textwidth}
        \centering
        \includegraphics[width=0.95\linewidth]{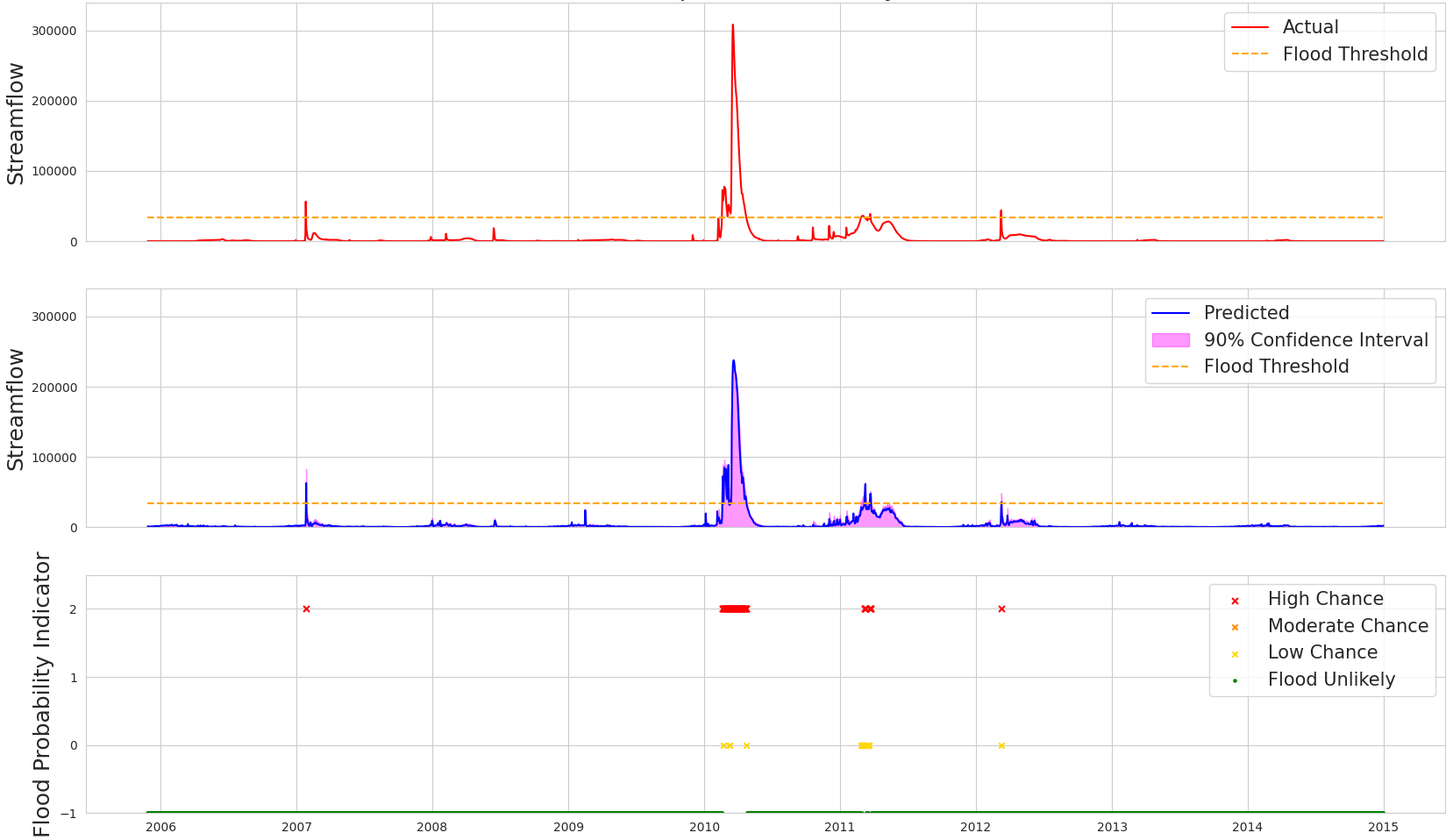}
        \caption{Cullyamurra Water Hole (Station ID: A0030501) in Cooper Creek of South Australia}
    \end{subfigure}
    \begin{subfigure}[b]{0.98\textwidth}
        \centering
        \includegraphics[width=0.95\linewidth]{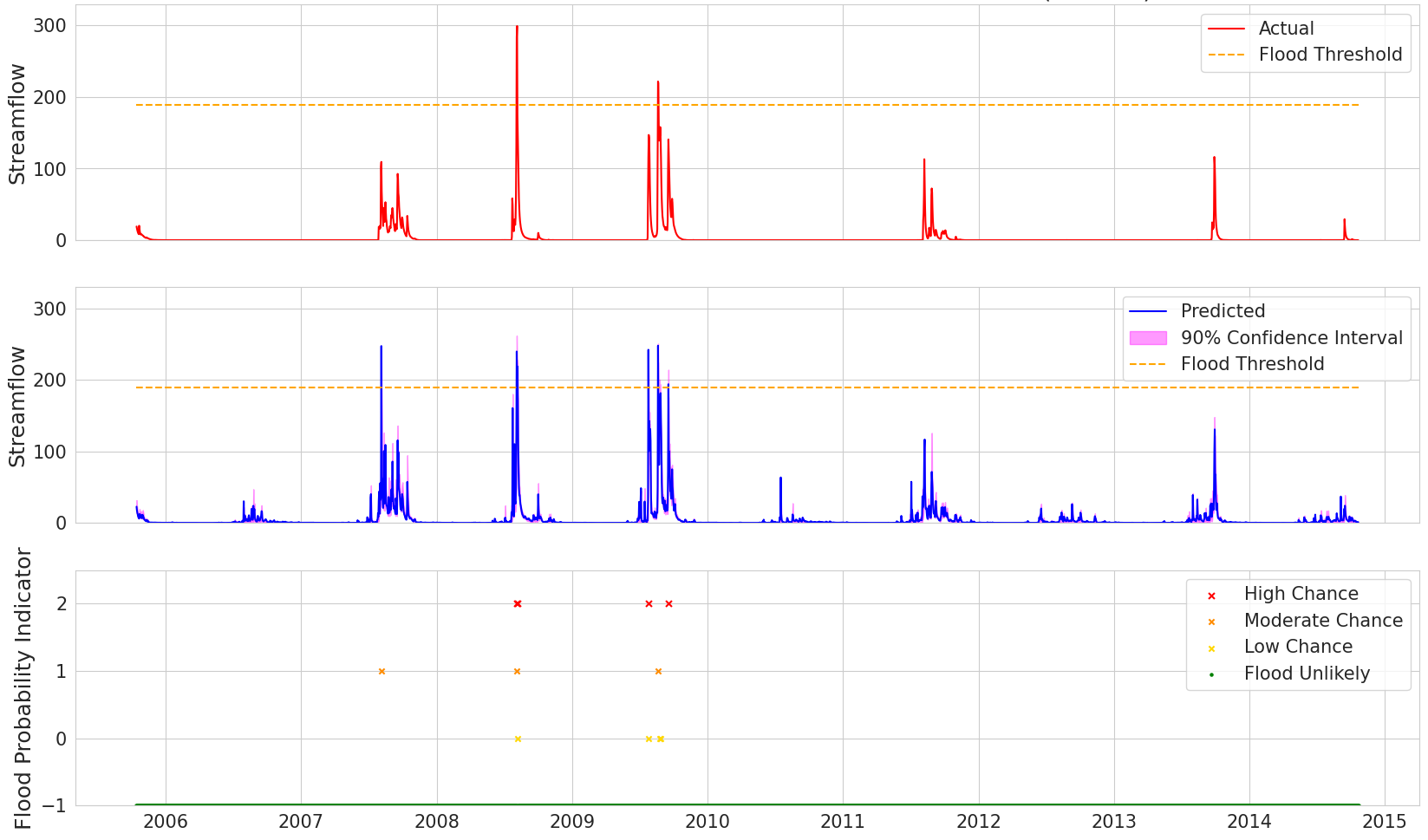}
        \caption{Darkin River at Pine Plantation (616002) in Swan coast - Avon River region of Western Australia}
    \end{subfigure}
    \caption{Flooding events in selected catchments of South Australia and Western Australia predicted using the Ensemble Quantile-LSTM streamflow prediction model }
    \label{fig:flood_pred_2}
\end{figure*}

 Finally, Figure \ref{fig:flood_pred_1} and \ref{fig:flood_pred_2} present the 5-day average of actual and predicted streamflow and associated flood risk indicators for 4 selected catchments in New South Wales, Queensland, South Australia, and Western Australia, respectively, using the Ensemble Quantile-LSTM model. The first catchment (\ref{fig:flood_pred_1}(b)), Barambah Creek River at Litzows, exhibits multiple flood events over the study period, with both actual and predicted streamflows exceeding the flood threshold on multiple occasions. The predicted streamflow uncertainty, represented by the 90\% confidence interval, captures most observed peaks, indicating the model's ability to anticipate extreme events. The associated flood risk indicator can capture the two major actual flooding events in early 2011 and 2013, respectively. Similarly, for Nepean River at Maguires Crossing, the model successfully identifies 3 of the 4 major flood events, albeit with some variability in prediction accuracy. Overall, these results demonstrate the model's capability to capture flood dynamics across diverse hydrological conditions, which is critical for early warning and disaster management. A similar performance is observed with Cullyamurra Water Hole (Figure \ref{fig:flood_pred_2}(a)) and Darkin River at Pine Plantation (Figure \ref{fig:flood_pred_2}(b)).


\section{Discussion}\label{disc}

In the first set of experiments, we compared the performance of the modelling strategies at the catchment scale using a conventional LSTM model for  five-step-ahead (daily) prediction of streamflow rate using climate data. We found that the Individual catchment modelling strategy, where a separate model is used for each gauging station data, provided  the best results (Table \ref{tab:catchment}). This is because each catchment has unique characteristics that could be based on weather patterns, climate,  and topography.   In Table \ref{tab:catchment},  we also observed that the Stacked Ensemble   model gives a poor performance, as it cannot beat the individual LSTM models. This could be due to a lack of information flow in combining the time series data with the static model input. \textcolor{black}We note that static data type is different when compared to time series data and there is a major challenge in machine learning when considering various types of data in data fusion strategies \cite{meng2020survey} and multiview learning \cite{zhao2017multi}. Comparatively,  \textcolor{black}{The Batch-Temporal strategy is essentially the Batch Model without the inclusion of static features. It operates on the same time-series features as the Individual strategy, but with a collective approach, running a single model for all stations in the state rather than individual models for each station. This consolidation aims to optimise computational resources.} The  Stacked Ensemble strategy obtained predictions from the temporal data using one model and then considered the static data in another model to more accurately capture temporal and static information in an ensemble. The Individual strategy features a specific fit to one catchment at a time and is unaffected by information from other catchments. For example, catchments might vary due to variations in weather patterns, climate, and topography.  Overall, these findings are consistent with the work of Leontjeva et al. \cite{leontjeva2016combining} who similarly compared a variety of strategies i) standalone model using univariate temporal data ii) standalone model using bimodal data (temporal and static) iii) ensemble model using bimodal data. In that study, the standalone model using bimodal data failed to capture information from both sources, while the performance of the ensemble model depended on the distribution of the data.

Generally, it is difficult for a single machine learning model to represent a wide range of data, even if they may be from the same domain, such as in a regionalisation context.
Regionalisation would be a challenge for machine learning in general, as a single model cannot be used (fitted to) for several tasks, and in this case, the local catchments can be seen as different tasks as they can have differences in the data distribution dependent on climate/weather/geography. Approaches such as multitask and transfer learning \cite{zhuang2020comprehensive,zhang2021survey} could be considered, where some underlying model knowledge can be re-used and applied to related datasets, i.e. catchments in this case.

In the case of deep learning models using the Individual catchment strategy (Table \ref{tab:dlmodels}),  the  Ensemble Quantile-LSTM model provides a mechanism for quantifying uncertainty with close prediction accuracy to conventional LSTM and CNN models. Although designed for image processing purposes, the CNN reproduces streamflow better than conventional LSTM models (Table 4) that were designed for temporal sequences. CNNs may also have advantages for capturing spatial regions (rationalisation) due to their capabilities in image processing, but LSTM models were the best due to their capacity to simulate temporal data. \textcolor{black}{However, Quantile LSTM (q=0.05) in Table 4 provides the best performance accuracy as it has been designed to capture extreme values.} The performance of these types of models has been comparable (close), especially for multistep ahead prediction  \cite{chandra2021evaluation}. Furthermore, the Individual catchment strategy does not incorporate the spatial catchment data into the deep learning models, although multivariate data of the local catchment was used. Hence, this could be another reason why LSTM models outperformed the CNN.   ED-LSTM and BD-LSTM  did not provide the best performance, contrary to past studies for univariate multistep ahead prediction \cite{chandra2021evaluation}. \textcolor{black}{However, we note that our study featured a multivariate approach with a much larger dataset, hence a direct comparison cannot be done.}

 The results, reinforce the efficacy of the  Ensemble Quantile-LSTM as a robust and adaptable choice for hydrological modelling, specifically for South Australia. This performance highlights the  Ensemble Quantile-LSTM model's resilience and effectiveness in capturing both extreme and regular streamflow which can imply their applicability in predicting flood occurrences.   The prediction of extreme values can potentially be corrected with data augmentation with deep learning models. We note that data augmentation and ensemble learning methods have been prominent \cite{khan2023review} in class imbalance problems, and we can view the case of extreme value forecasting as a regression case of class imbalance problems.


 \textcolor{black}{The predictions essentially give an outlook - up to five days ahead, which has been demonstrated to be useful for flood prediction based on historical flooding events for selected catchments. Although the model performance accuracy deteriorates as the prediction horizon increases in multistep ahead problems \cite{chandra2021evaluation}, it can still provide a forecast of upcoming flooding events and enable better disaster management. The results deteriorate as the gap in the values to predict increases with increasing prediction horizon. In an operational sense, we can envision a model that can be run repeatedly in real-time, providing more and more accurate predictions as more data becomes available in time. We have demonstrated that the streamflow prediction can be used as a means to enable forecasting of floods, as done in past studies \cite{kumar2023advanced,ha2021prediction,siqueira2020potential,zhong2023hybrid}. }
 
 Despite the success of evaluating several machine learning approaches for predicting flood extremes, this study has some obvious limitations. Firstly, the models do not naturally project model uncertainties into the predictions, which is vital for natural disasters caused by extreme events. 
 This can be addressed by Bayesian deep learning \cite{chandra2022revisiting,chandra2023bayesian} and variational deep learning models \cite{kapoor2023cyclone} in future work.  
 Secondly, we have not accounted for topography information in the data to develop a robust spatiotemporal model. 
 
  In the case of larger catchments, the steepness and variation in topography can influence rainfall patterns, particularly the within-catchment distribution of the rainfall. In addition, topography and catchment shape can influence the timing between rainfall and flood extremes. Implicitly, the proposed model should be able to capture these variations for the individual catchment strategy based on the data, but it is possible that adding additional information will assist the model in identifying the patterns.  We did not incorporate the changes in climate conditions due to the El Niño climate effect, which is a longer scale pattern in the Australian rainfall \cite{wang2007sensitivity,cai2009nina}. Again, the LSTM should be able to discover this from the data, given a long enough time series. However, adding this as a variable might assist the model in identifying the variations in the data.  
  The data is also sparse in space given the locations of the gauging stations (catchments), and this means if we need data for specific regions where no gauging stations are available, generative machine learning models \cite{harshvardhan2020comprehensive}, and general circulation models \cite{rocheta2014well} could be used.

\section{Conclusions}\label{conc}

 In this study, we undertook a critical evaluation of several deep learning models and presented a novel  Ensemble Quantile-LSTM model to accurately forecast streamflow extremes over selected Australian catchments using a multi-step-ahead prediction approach. Our investigation revealed that the Individual catchment strategy which features a separate model for a catchment gauging station gives the best performance for extreme value forecasting. Therefore, we found that a tailored approach, where each catchment is paired with a model dedicated to its distinct gauging station, significantly elevated the accuracy of extreme value forecasts. It's important to highlight that this precision in modelling ensures that local variability and unique hydrological characteristics are captured effectively, providing a more nuanced understanding of each catchment's behaviour in the face of extreme weather events.

The investigation unveiled the potential of our   Ensemble Quantile-LSTM model that integrates quantile regression to surpass all baseline machine learning models for streamflow prediction. This model emerges as a superior tool not only due to its performance but also because it represents a leap towards a more reliable prediction by embracing the inherent uncertainties in hydrological forecasting and addressing the problem of extreme value forecasting. Looking ahead, we recognise that the path to impeccable forecasts is iterative and progressive. Our research underscores the vast room for improvement that can be addressed by extending our dataset and refining our model, particularly in the area of uncertainty quantification. The aspiration is to reach a point where our model can offer robust predictions with quantified confidence levels, transforming how stakeholders prepare for and respond to the challenges posed by extreme rainfall and associated streamflow conditions such as floods.  \textcolor{black}{Our Ensemble Quantile framework can be improved further for uncertainty quantification by combining it with  Bayesian deep learning. }


 \section*{Software and Data Availability}

We provide meta-information for  open source code and data for the framework via Github repository:

 \begin{compactitem}
 
    \item[-] Name: Streamflow-floods
     \item[-] Developers: Siddharth Khedkar, Arpit Kapoor and Rohitash Chandra
    \item[-] Contact email: siddharthkhedkar11@gmail.com and c.rohitash@gmail.com
   \item[-] Compatible Operating System: Mac/Linux/Windows
    \item[-] Developed and tested: Ubuntu 20.04 (Linux)  
     \item[-] Year published: 2024
    \item[-] Source: Github   \footnote{\url{https://github.com/DARE-ML/streamflow-floods}}

\end{compactitem}

\section*{Author contributions}

A. Kapoor, S. Khedkar and J. Ng contributed to coding, writing and experiments. W. Vervoort contributed to analysis, editing and supervision. R. Chandra contributed to conceptualisation, writing, code analysis, and supervision. 

 \bibliographystyle{elsarticle-num}

\bibliography{cas-refs.bib}


\end{document}